\documentclass[runningheads]{llncs}
 
\usepackage[T1]{fontenc}
\usepackage{graphicx}
\usepackage{caption} 
\usepackage{subcaption}

\usepackage{amsmath}
\usepackage{amssymb}
\usepackage{booktabs}
\usepackage[table]{xcolor}
\usepackage{multirow}
\usepackage{etoolbox}
\usepackage{hyperref}
\addtocounter{tocdepth}{2}
\usepackage[capitalize,noabbrev]{cleveref}
\begin{document}

\title{Residuals Are Not Enough: Limits of Physics-Informed Pre-Training for Scientific Foundation Models}
\titlerunning{Physics-Informed Pre-Training for Scientific Foundation Models}
\author{
    Serge Kotchourko\inst{1} \and
    Amin Totounferoush\inst{1,2} \and \\
    Michael W. Mahoney\inst{2} \and
    Steffen Staab\inst{1,3}
}
\authorrunning{S. Kotchourko et al.}

\institute{
    University of Stuttgart, 70174 Stuttgart, Germany \email{serge.kotchourko@ipvs.uni-stuttgart.de, \{amin.totounferoush, steffen.staab\}@ki.uni-stuttgart.de} \and
    International Computer Science Institute, Lawrence Berkeley National Laboratory, University of California, Berkeley CA 94720, USA \email{mmahoney@stat.berkeley.edu} \and 
    Electronics and Computer Science,  University of Southampton, Southampton SO17~1BJ, United Kingdom
    \email{S.R.Staab@soton.ac.uk}
     }

\maketitle
\begingroup
\renewcommand{\thefootnote}{}%
\footnotetext{Equal contribution by Amin Totounferoush and Serge Kotchourko.}
\footnotetext{Code can be found on \href{https://github.com/kotserge/residuals-are-not-enough-limits-of-physics-informed-pre-training-for-scientific-foudation-models}{github.com}}
\endgroup

\begin{abstract}
Scientific foundation models (SciFMs) aim to learn generalizable representations of physical systems governed by partial differential equations (PDEs), enabling transfer across tasks and domains. While physics-informed methods, which leverage PDE residuals as supervisory signals, have shown promise in scientific machine learning (SciML) for improving accuracy and reducing data requirements, their potential in the context of SciFMs remains relatively unexplored.

In this evaluation study, we investigate whether (and how) physics-informed pre-training improves the generalization, robustness, and data efficiency of SciFMs. We conduct systematic experiments across a diverse set of PDEs, ranging from simple problems with periodic boundary conditions to more challenging systems such as the Navier-Stokes equations and non-periodic geometries.

Our results show that physics-informed pre-training provides clear benefits in ``nice,'' e.g., structured, well-aligned settings: it enhances generalization and reduces data dependence, compared to data-only pre-training. However, these advantages diminish significantly as the downstream tasks become ``harder,'' e.g., as they involve discontinuities or deviate from the pre-training distribution. In complex or structurally different problems, such as those involving new boundary conditions or PDE operators, physics-informed models may perform only on par with---or even worse---than data-driven baselines.

While residual-based pre-training helps in idealized regimes, realizing broadly transferable SciFMs will likely require subtler spatiotemporal inductive biases and more principled integration of physical knowledge into model architectures.

\end{abstract}
\section{Introduction}
\label{sec:intro}

By leveraging large-scale data and pre-training, as well as incorporating domain-specific inductive biases into architectures, foundation models have enabled the learning of representations that transfer to many downstream tasks with minimal effort, thereby transforming computer vision (CV)~\cite{dosovitskiy2021an,radford2021learning} and natural language processing (NLP)~\cite{achiam2023gpt,radford2019language}. 
Hoping to build on these advances in scientific machine learning (SciML), there is growing interest in scientific foundation models (SciFMs): broadly pre-trained, reusable models for families of problems governed by partial differential equations (PDEs) and other complex spatiotemporal processes, rather than models trained for a single equation, dataset, or task~\cite{shashank2023towards}. 

Such problems are central to SciML, where traditional numerical solvers, e.g., finite-difference, finite-element, and finite-volume methods, are computationally expensive, scale poorly with complexity, and/or require expert tuning, motivating faster and more generalizable alternatives.
Physics-informed neural networks (PINNs)~\cite{karniadakis2021physics,raissi2019physics,krishnapriyan2021characterizing} try to address this by incorporating physical constraints into the learning process to enforce compliance with the governing equations.
Nevertheless, they are brittle, and they generalize poorly to new parameter distributions or PDE types without extensive retraining~\cite{krishnapriyan2021characterizing}. 
Similarly, neural operators and Fourier neural operators (FNOs)~\cite{kovachki2023neural,li2020fourier,FalsePromizeZeroShot_TR} learn mappings between function spaces, improving adaptability across PDE parameters.
Nevertheless, they remain tied to the operators seen during training, and they require fine-tuning to handle new operators and/or to perform tasks such as zero-shot super-resolution~\cite{FalsePromizeZeroShot_TR}. 

SciFMs aim to generalize across many operators, including unseen ones, by pre-training on large, diverse PDE solution datasets and adapting to downstream tasks via fine-tuning~\cite{shashank2023towards} or in-context learning~\cite{yang2023context}.
A key limitation of current SciFM efforts is their heavy reliance on PDE solution data. 
Generalizing across many operators would require an extensive dataset spanning both a wide range of operators and a broad distribution of PDE parameters. 
This is very demanding: each operator has distinct mathematical structure, and even slight parameter shifts can drastically alter solution stability and behavior. 
For instance, increasing the advection-to-diffusion ratio in an advection-diffusion equation can induce severe numerical instability. 
Although successful SciML models, e.g., large weather models such as~\cite{bodnar2025foundation}, show that large-scale scientific pre-training can be highly effective when sufficiently rich datasets are available, for broader PDE-governed problems, assembling datasets of comparable scale and diversity remains computationally expensive and relatively underexplored, creating a key bottleneck for SciFM development.
This has motivated the development of ``neural digital twin'' models such as Neural Discrete Equilibrium (NeurDE)~\cite{benitez2025neural}.

To address the challenge of data scarcity in developing SciFMs, we investigate whether (and, if so, how) physics-informed pre-training---by incorporating PDE residuals into the training loss---can enhance downstream performance and reduce the need for labeled solution data. 
While physics-informed learning has been widely-explored in SciML~\cite{raissi2019physics,pino}, its effectiveness as a pre-training strategy for SciFMs remains largely unexplored.
Our study includes eight prototypical PDE systems commonly encountered in scientific and engineering applications. 
Our pre-training is conducted on three equations: Poisson, Advection-Diffusion, and Helmholtz. For evaluation, we consider five additional systems that introduce greater complexity and diversity: Reaction-Diffusion, Reaction-Advection-Diffusion, Darcy flow, and both laminar and turbulent Navier-Stokes equations.

\paragraph{Contributions and Scope of Investigation.}
\label{contributions}

\begin{figure}[t]
    \centering
    \includegraphics[width=0.55\linewidth]{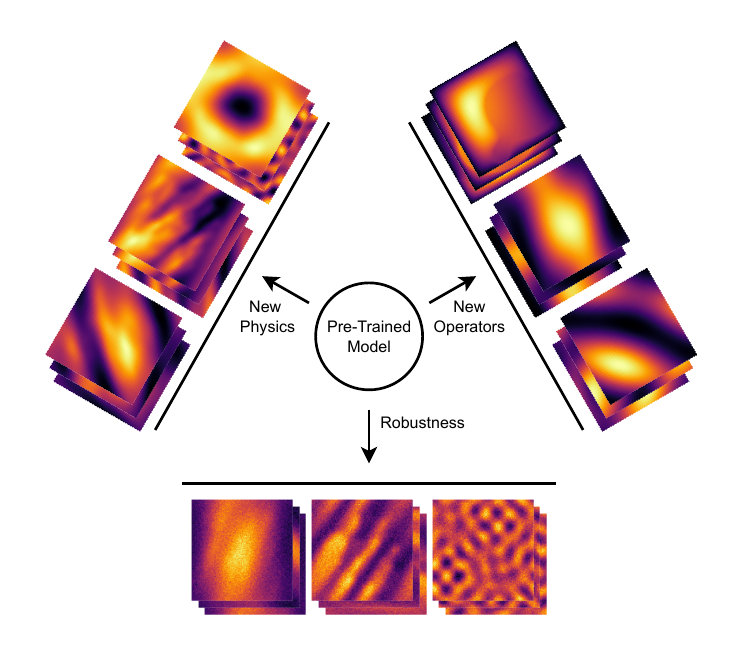}    
    \caption{We evaluate physics-informed pre-trained models on their ability to generalize to shifted parameter regimes, unseen PDE operators, and noisy fine-tuning data—capturing challenges in new physics, new operators, and robustness.}
    \label{fig:overview-contribution}
\end{figure}

We assess whether physics-informed pre-training can mitigate data scarcity, improve generalization, and increase robustness in noisy settings for SciFMs by evaluating model performance across key challenges, a setup which is shown in Figure~\ref{fig:overview-contribution}:
\begin{enumerate}
    \item \textbf{Generalization to new physics.} 
    We evaluate the model's ability to generalize within the same PDE operator under shifted PDE parameters. For instance, we pre-train the model with a diffusion coefficient sampled from \( D \sim (1, 5) \) and fine-tune it with \( D \sim (12, 15) \). This shift introduces significantly different behavior (physics) compared to the pre-training data.
    \item \textbf{Generalization to new operators.} 
    We assess generalization beyond the pre-trained operators $\mathcal{G}_1, \mathcal{G}_2, \dots, \mathcal{G}_n$ to entirely new ones, $\mathcal{G}_{\text{new-1}}, \mathcal{G}_{\text{new-2}}, \dots$, including the transition from steady-state PDEs with periodic boundaries to time-dependent PDEs with Dirichlet or Neumann boundaries.
    \item \textbf{Robustness to noisy fine-tuning data.}
    In practical scenarios, fine-tuning data is often corrupted by noise. We analyze the robustness of the physics-informed foundation model when fine-tuned with noisy data.
    Our goal is to determine the extent to which physics-informed learning aids in mitigating the impact of noise and enhances model robustness.
\end{enumerate}
Overall, our results indicate that, while existing pre-training methods perform reasonably-well for relatively-``nice'' settings, advantages diminish when downstream tasks become more challenging.
Taken together, our investigations clarify how physics-informed pre-training might (or might not) mitigate data scarcity and computational expense in SciFM development, along with pointing out its limitations and important directions for improvement.
These future directions include the need for improved neural surrogate models~\cite{benitez2025neural} as well as for improved understanding of the implicit effects of model design choices~\cite{annan2_understanding_biasses_TR}.
\section{Related Works}

Two popular categories of machine learning approaches to PDE problems include PINNs~\cite{raissi2019physics,karniadakis2021physics,krishnapriyan2021characterizing}, which enforce physical consistency through loss constraints, and neural operators such as FNOs~\cite{li2020fourier,pino} and DeepONets~\cite{lu2021learning,goswami2022physics}, which learn mappings between function spaces.
However, such approaches face fundamental challenges with new physics, new geometries, etc.~\cite{krishnapriyan2021characterizing,FalsePromizeZeroShot_TR}, and thus they tend to generalize poorly beyond the parameters and operators seen during training.

Transfer learning has emerged as a potential approach to address these limitations, enabling models to adapt to new problems with reduced computational overhead. 
Previous studies have explored transfer learning between different geometries, resolutions and scales~\cite{xu2023transferload,wang2022mosaic,song2022transfer,ho2024probabilistic}, demonstrating the ability of these models to generalize under very specific conditions. 

Recent work has focused on improving generalization and efficiency in solving parametric PDEs. 
For example, \cite{cho2023hypernetwork} propose low-rank models with a hypernetwork-based meta-learning framework for fast approximation across varying inputs; \cite{huang2022meta} introduce MAD, a mesh-free, unsupervised method that encodes PDE parameters as latent vectors for rapid adaptation without retraining; and \cite{kassai2024boosting} present GEPS, a low-rank adaptive conditioning mechanism that enhances generalization to unseen conditions through first-order updates on context parameters. 

Another line of work incorporates physical knowledge directly into the architecture of neural networks, embedding structure beyond loss-level constraints. Most notably, the Neural Discrete Equilibrium (NeurDE) model~\cite{benitez2025neural} is a ``neural digital twin'' that integrates a kinetic-theory ``lifting'' into its design, learning equilibrium operators within a lattice Boltzmann framework, thereby enabling accurate long-term flow forecasting with physically grounded representations.

A new line of work has also explored in-context learning for PDEs~\cite{yang2023context,chen2024data}, in which transformer-based models leverage demonstrations provided in the prompt to solve differential equations. These models map input queries to correct solutions by mimicking the examples in the prompt, enabling them to address tasks similar to those seen in the demos. 
While promising, this approach heavily relies on the quality and diversity of the training demos, and its ability to generalize to new problem settings, such as unseen PDEs, remains a significant challenge.

Another line of research focuses on pre-training and fine-tuning strategies, as explored in~\cite{shashank2023towards,mccabe2024multiple,rahman2024pretraining,herde2024poseidon}. The work most closely related to ours is~\cite{shashank2023towards}, which investigates transfer learning across PDE operators within a pre-training and fine-tuning framework. 
Their study examines factors such as the impact of fine-tuning dataset size, model scale, and generalization to the same PDE with different parameter distributions, as well as to entirely new PDE types. 
Subsequent work~\cite{mccabe2024multiple} substantially reproduced the results of~\cite{shashank2023towards}.

These studies highlight the potential of pre-training and fine-tuning for improving generalization in SciML. However, they all rely on supervised pre-training with solution data, which limits scalability when hoping to develop SciFMs across diverse PDEs. In contrast, our work investigates whether physics-informed pre-training can serve as a more data-efficient alternative. Our findings clarify the strengths and limitations of this approach, offering insight into the viability of physics-informed pre-training for building broadly applicable SciFMs.

\section{Methodology}
\label{sec:method}

We aim to approximate solutions to PDEs using machine learning techniques. Mathematically, a general form of a PDE can be expressed as:
\begin{equation}
\mathcal{G}(u; \lambda) = f(\mathbf{x, t}), \quad \mathbf{x} \in \Omega, \quad \mathbf{t} \in \tau,
\label{equ:operator}
\end{equation}
where \( \mathcal{G} \) is a differential operator (e.g., Laplace, Poisson, or Navier-Stokes operator), \( u(\mathbf{x}) \) is the unknown solution we seek to approximate, \( \lambda \) represents the PDE parameters such as material properties, \( f(\mathbf{x, t}) \) is the force function, \( \Omega \) is the spatial domain, and \(\tau \) is the time domain.

Our objective is to learn a model \( \mathcal{F_{\theta}} \), which approximates the solution \( u(\mathbf{x, t}) \) by mapping the force function, initial condition, and PDE parameters to the solution:
\begin{equation}
\mathcal{F_{\theta}}(f, \lambda) \approx u(\mathbf{x, t}).
\end{equation}
where $\theta$ describes the model's parameters.\footnote{In this work, all model inputs and outputs are discrete representations of the underlying continuous functions; issues associated with this discretization step are non-trivial and have been highlighted previously~\cite{krishnapriyan2021characterizing,differentiable_hard_22_TR,continuous22_TR,learnConservation1_TR,benitez2025neural,FalsePromizeZeroShot_TR}.}
We investigate a two-stage learning strategy that consists of pre-training on a diverse set of PDEs, followed by fine-tuning on specific tasks, to assess the effectiveness of physics-informed pre-training for SciFMs.

\subsection{Training and Evaluation PDE Systems Setup}
\label{pdes}
We consider eight fundamental PDE systems for both pre-training and evaluation. The coefficients are detailed in Table~\ref{tab:pdes}. Pre-training is conducted on three steady-state equations~\cite{shashank2023towards}; Poisson, Advection-Diffusion, and Helmholtz. For these PDEs, we evaluate generalization across both in-distribution (ID) and out-of-distribution (OOD) parameter regimes. Evaluation further includes five entirely new PDEs: Reaction-Diffusion, Advection-Reaction-Diffusion, Darcy flow~\cite{takamoto2022pdebench}, and two variants of the Navier-Stokes equations—laminar and turbulent. All problems are defined on \( \Omega = [0,1]^2 \), with time-dependent PDEs evolving over \( \tau = [0,1] \). Periodic boundary conditions are used except for Darcy, which has Dirichlet boundaries. Steady-state problems learn mappings from force and parameters to the solution; while time-dependent problems map parameters and initial conditions to the solution at \( t = 1 \).
\begin{table}
    \centering
    \caption{Scalar PDE coefficient ranges \((\min, \max)\) for pre-training and fine-tuning: diffusion \( D \), advection-diffusion ratio \( \Psi \), wave number \( \omega \), reaction term \( r \), velocity field \( v \), kinematic viscosity \( \nu \).
    }
    \resizebox{\linewidth}{!}{%
        \begin{tabular}{@{}lllllll@{}}
\toprule
\multirow{2}{*}{Training-Stage} & \multirow{2}{*}{Benchmark}     & \multirow{2}{*}{System}                    & \multicolumn{4}{l}{Coefficient Range}                                                               \\
                                &                                &                                            & ID              & Slight-OOD      & Medium-OOD      & High-OOD           \\ \midrule
\multirow{3}{*}{Pre-Training}   &  \multirow{3}{*}{-}            & Poisson (\( D \))                          & {$(1, 5)$} & \multicolumn{3}{c}{-}\\
                                &                                & Advection-Diffusion       (\( \Psi \))     & {$(0.2, 1)$} & \multicolumn{3}{c}{-} \\
                                &                                & Helmholtz                 (\( \omega \))   & {$(1, 10)$} & \multicolumn{3}{c}{-} \\ \midrule
\multirow{8}{*}{Fine-Tuning}     & \multirow{3}{*}{New Physics (zero- \& n-shot)}   & Poisson (\( D \))                                   & $(1, 2.5)$ & $(2.5, 7.5)$ & $(7.5, 12.5)$ & $(15, 20)$ \\
                                &                                & Advection-Diffusion       (\( \Psi \))                & $(0.2, 0.4)$     & $(0.4, 1.6)$       & $(2, 3)$            & $(4, 5)$         \\
                                &                                & Helmholtz                 (\( \omega \))                 & $(1, 5)$       & $(2, 12)$        & $(10, 13)$        & $(12, 15)$     \\ \cmidrule{2-7} 
                                & \multirow{2}{*}{Robustness}    & Advection-Diffusion       (\( \Psi \))                  &                            & $(0.4, 1.6)$       & $(2, 3)$            & $(4, 5)$         \\
                                &                                & Helmholtz                 (\( \omega \))                 &                            & $(2, 12)$        & $(10, 13)$        & $(12, 15)$     \\ \cmidrule{2-7} 
                                & \multirow{4}{*}{New Operators} & Reaction-Diffusion                         & \multicolumn{4}{c}{$\mathbf{D} \sim (5, 10), r \sim (-1, 1)$}                                                          \\
                                &                                & Reaction-Advection-Diffusion               & \multicolumn{4}{c}{$\mathbf{D} \sim (5, 10),\mathbf{v} \sim (0.1, 1), r \sim (-1, 1)$}                                 \\
                                &                                & Navier-Stokes                              & \multicolumn{4}{c}{\( \nu \sim (1e-5, 1e-3)\)}                                                                         \\
                                &                                & Darcy                                      & \multicolumn{4}{c}{-}                                                                         \\ \bottomrule
\end{tabular}%
    }
    \label{tab:pdes}
\end{table}

\subsection{Pre-training Strategies}
To assess the impact of physics-informed pre-training, we compare it against a purely data-driven counterpart.
\subsubsection{Pre-training with Data Loss (baseline)}
In the purely data-driven setting, the model is pre-trained to minimize the discrepancy between predicted and ground-truth solutions. The loss function is defined as:
\begin{equation}
    \mathcal{L}_{\text{data}}(\theta) = \frac{1}{M} \sum_{j=1}^{M} \left\| \mathcal{F}_{\theta}(f_j, \lambda_j) - u_j \right\|^2,
\end{equation}
where \( \mathcal{F}_{\theta} \) denotes the neural network parameterized by \( \theta \), \( f_j \) is the forcing term, \( \lambda_j \) are PDE parameters, and \( u_j \) is the ground-truth solution.

\subsubsection{Physics-Informed Pre-training (investigated approach)}
This pre-training strategy combines data loss with physics-based constraints in the form of PDE residuals. The hybrid loss is defined as:
\begin{equation}
    \mathcal{L}_{\text{hybrid}}(\theta) = \alpha \mathcal{L}_{\text{data}}(\theta) + (1 - \alpha) \mathcal{L}_{\text{PDE}}(\theta),
\end{equation}
where:
\begin{equation}
    \mathcal{L}_{\text{PDE}}(\theta) = \frac{1}{N} \sum_{i=1}^{N} \left\| \mathcal{G}(\mathcal{F}_{\theta}(f_i, \lambda_i)) - f_i \right\|^2,
\end{equation}
and \( \mathcal{G} \) is the PDE operator applied to the predicted solution. The coefficient \( \alpha \in [0,1] \) balances the influence of data and physics terms. This formulation is similar to the approach presented in~\cite{pino}. We use spectral derivatives to compute PDE residuals for problems with periodic boundary conditions, and finite-difference approximations for non-periodic cases.
\subsection{Fine-tuning Strategy}
After the pre-training phase, the model is fine-tuned using only data loss to adapt to specific tasks and improve accuracy for target PDE instances. Fine-tuning refines learned representations and optimizes performance for downstream tasks, including PDEs with shifted parameter distributions and unseen PDEs.

Given a dataset \( \mathcal{D}_{\text{fine}} = \{ (f_j, u^0_j, \lambda_j, u_j) \}_{j=1}^{M} \), where \( f_j \) represents the forcing function, $u^0_j$ is the initial condition (in the case of time-dependent problems), \( \lambda_j \) denotes the PDE parameters, and \( u_j \) is the corresponding ground-truth solution, the fine-tuning process minimizes the following data-driven loss:
\begin{equation}
    \mathcal{L}_{\text{data}}(\theta) = \frac{1}{M} \sum_{j=1}^{M} \left\| \mathcal{F}_{\theta}(f_j, u^0_j, \lambda_j) - u_j \right\|^2,
\end{equation}
where \( \mathcal{F}_{\theta} \) represents the neural network parameterized by \( \theta \). 
Fine-tuning with data loss allows the model to adapt efficiently to specific PDE instances, leveraging the representations learned during pre-training while focusing on task-specific accuracy.

\section{Experimental Setup}
Dataset generation, training, inference, and implementation follow established practices introduced previously~\cite{shashank2023towards}. 

\subsection{Data Augmentation} 
To evaluate the robustness of pre-trained models during fine-tuning, we simulate real-world measurement noise by perturbing the solution data with additive white Gaussian noise. Specifically, noise is added as follows:
\begin{equation} 
    u_{\text{noisy}} = u + \sigma \cdot \text{std}(u) \cdot \epsilon, 
\end{equation} 
where \( \epsilon \sim \mathcal{N}(0, 1) \) is a standard normal random variable, and \( \sigma \) controls the noise level. We consider several levels of noise: \( \sigma = 0.01 \) to reflect high-precision measurements, \( \sigma = 0.05 \) and \( \sigma = 0.1 \) for moderate uncertainty, and \( \sigma = 0.2 \) to represent severe noise. This setup allows us to systematically examine the model’s sensitivity to varying degrees of noise during fine-tuning.
\subsection{Model Selection}
\label{subsec:model_selection}
We compare two architectures for physics-informed pre-training: FNO~\cite{li2020fourier} and the Axial Vision Transformer (AViT)~\cite{mccabe2024multiple}, both pre-trained with our framework and evaluated zero-shot on downstream PDE tasks. FNO consistently outperformed AViT: mean relative $\ell_2$ error ($\mu_{\ell_2}$, FNO vs.\ AViT) was $0.0005$ vs.\ $0.2852$ on Poisson, $0.0025$ vs.\ $0.3166$ on Advection-Diffusion, and $0.0007$ vs.\ $0.7792$ on Helmholtz, with the same ordering under slight distribution shift. Given this and its lower computational cost, we use FNO in most subsequent experiments.
\subsection{Baselines}
\label{subsec:baselines}
We compare the performance of physics-informed pre-trained models against two baseline variants. Both baselines use the FNO architecture and the downstream data loss \( \mathcal{L}_{\text{data}} \): \textbf{Scratch} is trained directly on each downstream task, and \textbf{Data-Loss} is pre-trained with \( \mathcal{L}_{\text{data}} \) alone~\cite{shashank2023towards} before fine-tuning.

\subsection{Evaluation Metrics}
\label{subsec:evaluation_metrics}
To evaluate the performance of the models, we employ two distinct metrics: the \( \mu_{\ell_2} \) error, the \( L_{\infty} \) error, and the fRMSE. The \( \mu_{\ell_2} \) error quantifies the mean relative error between the predicted and ground truth solutions, defined as 
\begin{equation}
    \mu_{\ell_2} = \frac{1}{N} \sum_{i=1}^{N} \frac{\| u_i - \hat{u}_i \|_2}{\| u_i \|_2},
\end{equation}
which captures the average deviation from the ground truth. 

The fRMSE describes the root mean square error in the frequency domain, defined as 
\begin{equation}
    \text{fRMSE} = \frac{1}{N} \sum_{i=1}^{N} 
    \frac{\sqrt{\sum_{k_{\min}}^{k_{\max}} \left| \mathcal{F}(\hat{u}_i) - \mathcal{F}(u_i) \right|^2}}{k_{\max} - k_{\min} + 1},
\end{equation}
where \( \mathcal{F} \) denotes the Fourier transform, and \( k_{\max} \) and \( k_{\min} \) are the maximum and minimum indices of the Fourier transform, respectively, with \( N \) representing the number of samples. We adopt a partitioning approach similar to that in \cite{takamoto2022pdebench}  for the classification of low, middle, and high-frequency regions. This classification allows us to quantify the model's ability to capture global, large-scale features, medium-scale features, and periodicity, as well as small-scale features and noise corresponding to the low, middle, and high-frequency regions, respectively.

\section{Empirical Results}
\label{sec:results}
We evaluate physics-informed pre-training across three scenarios: zero-shot generalization to ID/OOD tasks, \( n \)-shot learning with limited and noisy data, and transfer to unseen PDEs. This assesses the model’s ability to generalize, adapt to shifts, and remain robust under noise.
\subsection{Zero-Shot Generalization}
\label{subsec:zero-shot-generalization}
We first evaluate whether physics-informed pre-training improves zero-shot generalization on the same PDE operators used during pre-training, i.e., Poisson, Advection-Diffusion, and Helmholtz, under in-distribution (ID) and progressively out-of-distribution (Slight-, Medium-, and High-OOD) coefficient ranges. 
Parameter ranges are summarized in Table~\ref{tab:pdes}, and zero-shot results for the $\mu_{\ell_2}$ metrics are reported in Table~\ref{tab:zero-shot}. To isolate the contribution of pre-training, we additionally compare against a model trained from Scratch.
\begin{table*}
    \centering
    \caption{Zero-shot evaluation on downstream tasks governed by the pre-training PDEs, under in-distribution and slight, medium, and high out-of-distribution (S/M/H-OOD) parameterizations. Best per metric in bold.}
    \resizebox{\linewidth}{!}{%
        \begin{tabular}{llllllllllllll}
        \toprule
        \multirow{2}{*}{Metric} & \multirow{2}{*}{Model} & \multicolumn{4}{l}{Poisson} & \multicolumn{4}{l}{Advection-Diffusion} & \multicolumn{4}{l}{Helmholtz} \\
                & & ID & S-OOD & M-OOD & H-OOD & ID & S-OOD & M-OOD & H-OOD & ID & S-OOD & M-OOD & H-OOD \\ \midrule
        \multirow{3}{*}{$\mu_{\ell_2}$}
            & Scratch          & 1.0080          & 1.1736          & 1.0061          & 1.8586          & 1.0005          & 1.1850          & 1.2476          & \textbf{1.3294} & 1.0137          & 1.1039          & 1.5014          & 1.3249          \\
            & Data-Loss        & 0.0025          & 0.0102          & \textbf{0.1225} & \textbf{0.5750} & 0.0068          & 0.0434          & \textbf{0.8813} & 1.8948          & 0.0040          & 0.2448          & \textbf{1.2165} & 1.2121          \\
            & Physics-Informed & \textbf{0.0005} & \textbf{0.0067} & 0.1326          & 0.7091          & \textbf{0.0025} & \textbf{0.0260} & 1.2705          & 4.1350          & \textbf{0.0007} & \textbf{0.2363} & 1.2177          & \textbf{1.1825} \\ 
              \bottomrule
        \end{tabular}%
    }
    \label{tab:zero-shot}
\end{table*}

\paragraph{Pre-training helps in- and near-distribution.}
Both pre-trained models substantially outperform the Scratch baseline in the ID and Slight-OOD settings, and the Data-Loss results are consistent with the findings of~\cite{shashank2023towards}. The Physics-Informed model is particularly effective in the Helmholtz ID and Slight-OOD settings relative to Data-Loss, suggesting that incorporating PDE residuals as a training signal helps the model capture the underlying physical structure of the operator.

\paragraph{Sensitivity to large distribution shifts.}
These gains do \emph{not} persist as the coefficient ranges move further OOD. In the Medium- and High-OOD regimes, the ``physics-informed'' model degrades sharply, often falling below the Data-Loss baseline. 
This is consistent with prior work~\cite{krishnapriyan2021characterizing,FalsePromizeZeroShot_TR,benitez2025neural}.
The Helmholtz task makes this limitation clear: small changes in the wave number drastically alter the solution space, and a similar effect arises in
Advection-Diffusion, where large coefficient shifts introduce high velocities changing the solution significantly. Under these conditions the $\mathcal{L}_\text{PDE}$ signal acts as a liability rather than a useful regularizer, and the Data-Loss model proves more robust to coefficient variation. This points to overfitting to the pre-training distribution rather than the acquisition of transferable physical principles.

\subsection{$n$-Shot Generalization and Noise Robustness}
\label{subsec:n-shot-learning}
\begin{figure*}
    \centering
    \includegraphics[width=0.9\linewidth, trim=0 1cm 0 0, clip]{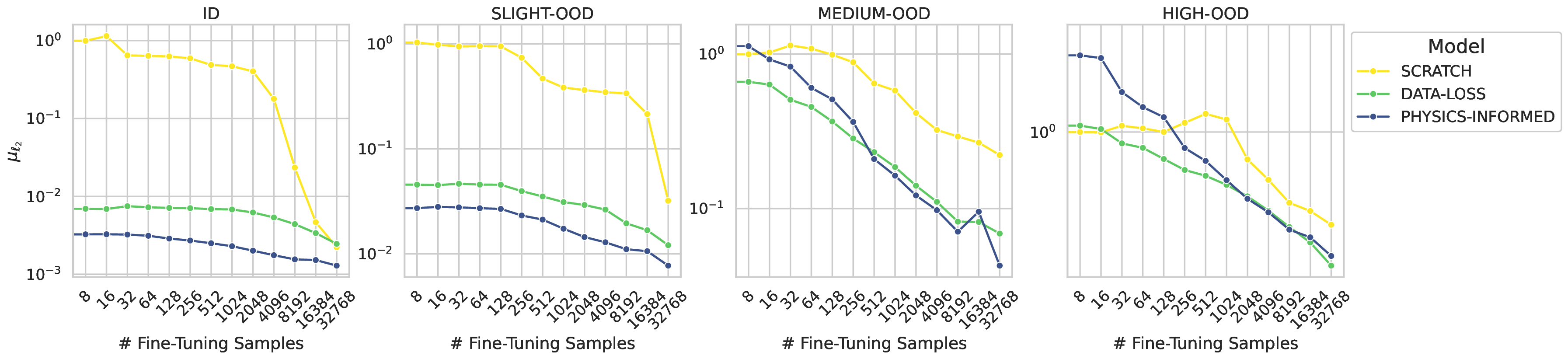}
    \includegraphics[width=0.9\linewidth, trim=0 0 0 1cm, clip]{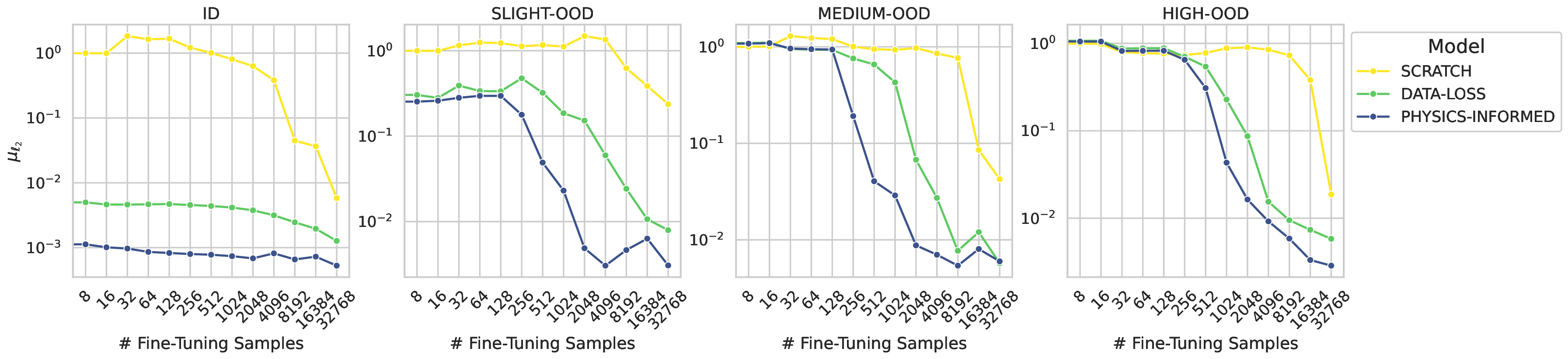}
    \caption{$\mu_{\ell_2}$ error of fine-tuned models on the Advection-Diffusion (top) and Helmholtz (bottom) tasks using up to $2^{15}$ downstream examples, across ID and OOD parameter regimes, comparing pre-training strategies.}
    \label{fig:fine-tuned}
\end{figure*}

\begin{figure*}
    \centering
    \includegraphics[width=.9\linewidth]{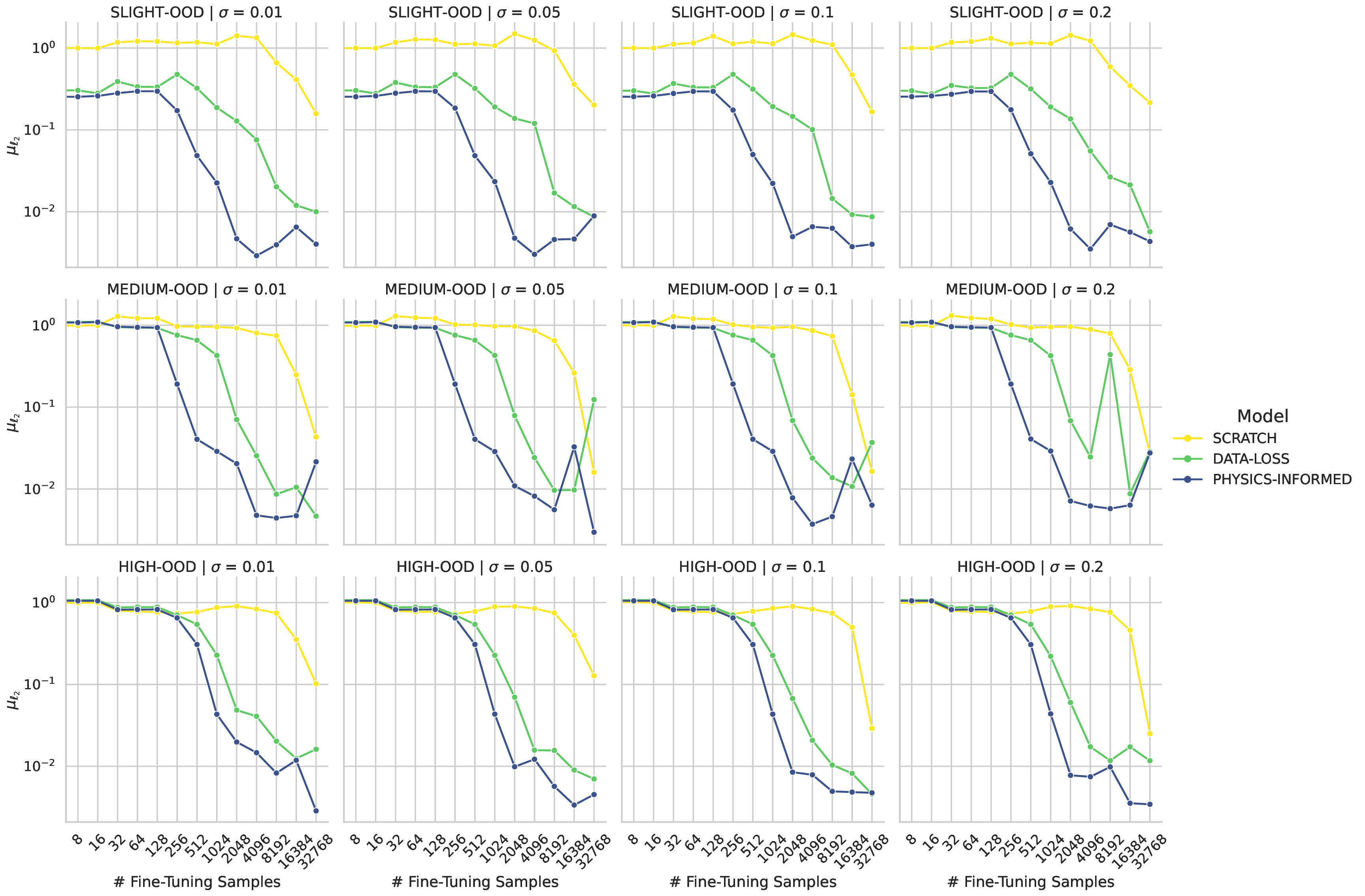}
    \caption{$\mu_{\ell_2}$ error for varying noise levels in the solution domain on the
    OOD Helmholtz datasets. From left to right, $\sigma = 0.01, 0.05, 0.1, 0.2$.}
    \label{fig:n-shot-base-and-pushed-helm-noise-mul2}
\end{figure*}

To assess $n$-shot generalization, we fine-tune the pre-trained models on downstream tasks with varying distribution shifts, using up to $2^{15}$ examples (Figure~\ref{fig:fine-tuned}). Each example corresponds to one complete input-output pair, consisting of a problem instance and its associated solution.
\paragraph{Generalization to new physics.}
Both pre-training strategies outperform training from scratch, and physics-informed models show clear gains in the ID and Slight-OOD settings, reaching comparable accuracy with fewer samples than the Scratch model (e.g., up to $8\times$ fewer on Helmholtz). 
This diminishes under larger shifts: in the Medium- and High-OOD regimes, and especially for Advection-Diffusion, the physics-informed advantage in low-data conditions disappears and can fall below the data-only baseline, recovering only at higher sample counts. 
The fRMSE analysis attributes this to spectral overfitting: physics-informed pre-training biases the model toward the frequencies that dominate the pre-training data. Performance degrades out-of-distribution precisely when a shift changes the solution to frequencies under-represented during pre-training. As a result, the model handles familiar frequencies well but adapts poorly to new spectral patterns.
\paragraph{Robustness under noisy fine-tuning.}
To assess robustness, noise is added to the downstream solution data, increasing the level from $\sigma = 0.01$ to $\sigma = 0.2$ across the OOD settings (Figure~\ref{fig:n-shot-base-and-pushed-helm-noise-mul2}).
Physics-informed models exhibit substantial robustness, as their accuracy is largely unaffected by noise, and outperform the data-only baselines at all noise levels. This advantage is most pronounced for the Helmholtz setting, where the improvement observed in the ID and Slight-OOD regimes is preserved even under strong perturbations.
The fRMSE breakdown clarifies this mechanism. Additive Gaussian noise primarily corrupts high-frequency modes, and physics-informed models retain consistently lower high-frequency errors, even when fine-tuned on noisy data, mirroring their noise-free behavior. The spectral alignment with the governing PDEs that limit out-of-distribution transfer therefore grants an advantage under noise: by constraining the model toward the true high-frequency content of the solution, the residual loss $\mathcal{L}_{\text{PDE}}$ enables the recovery of fine-scale features that purely data-driven training fails to capture.
\subsection{Shifting Focus to Unseen PDEs}
\label{subsec:new-pdes}
\begin{figure}
    \centering
    \includegraphics[width=0.88\linewidth]{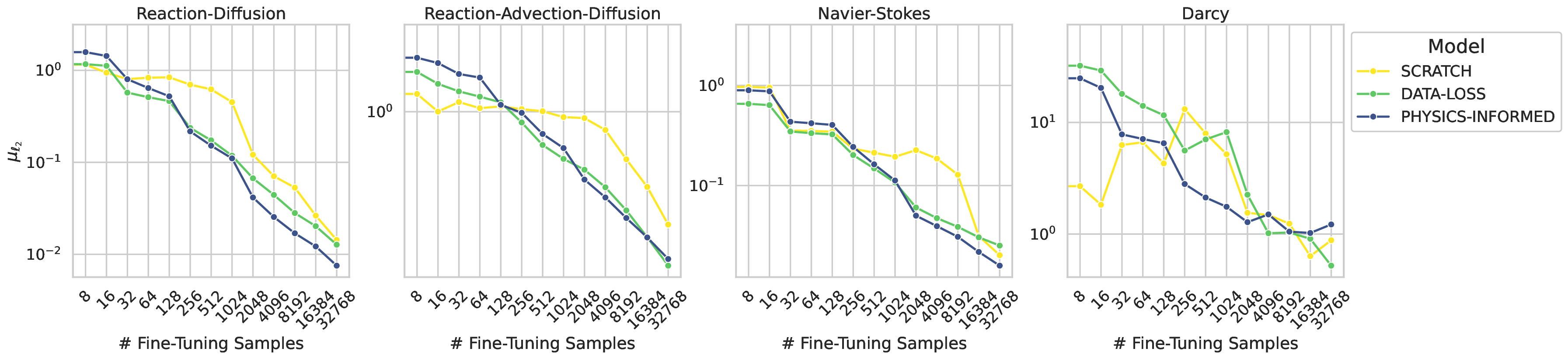}
    \caption{$\mu_{\ell_2}$ metrics with increasing numbers of examples used during fine-tuning for the downstream fine-tuning tasks.}
    \label{fig:new-operators-extended-dataset-mul2}
\end{figure}
\begin{figure}
  \centering
  \begin{subfigure}[b]{0.21\linewidth}
    \includegraphics[width=\linewidth]{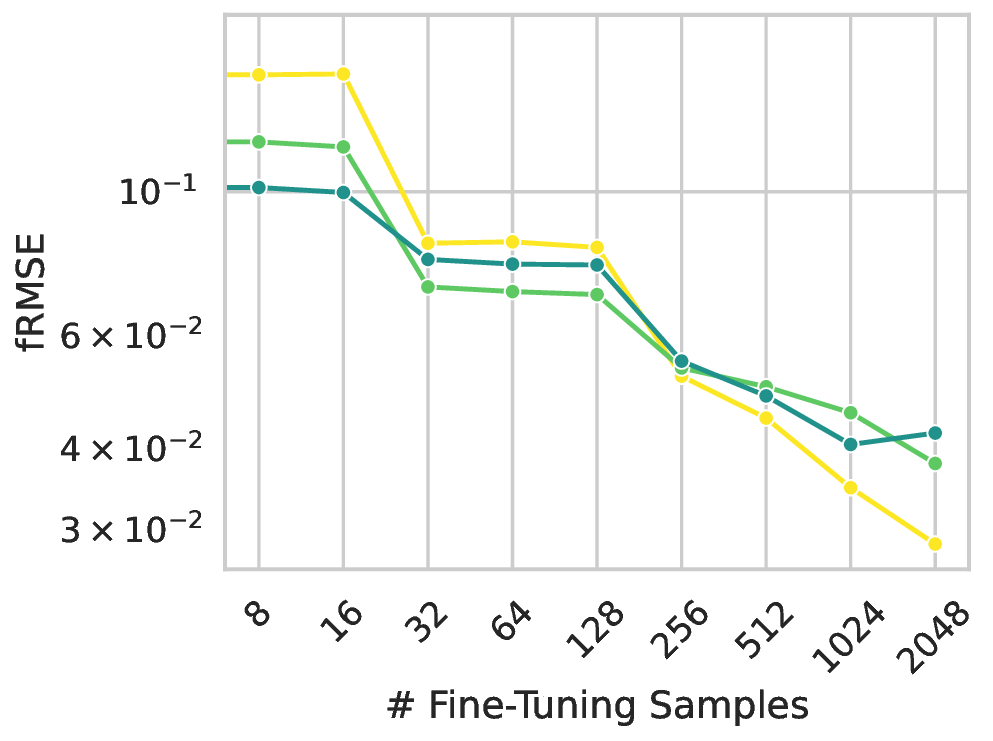}
    \caption{Low}
    \label{fig:low}
  \end{subfigure}\hfill
  \begin{subfigure}[b]{0.22\linewidth}
    \includegraphics[width=\linewidth]{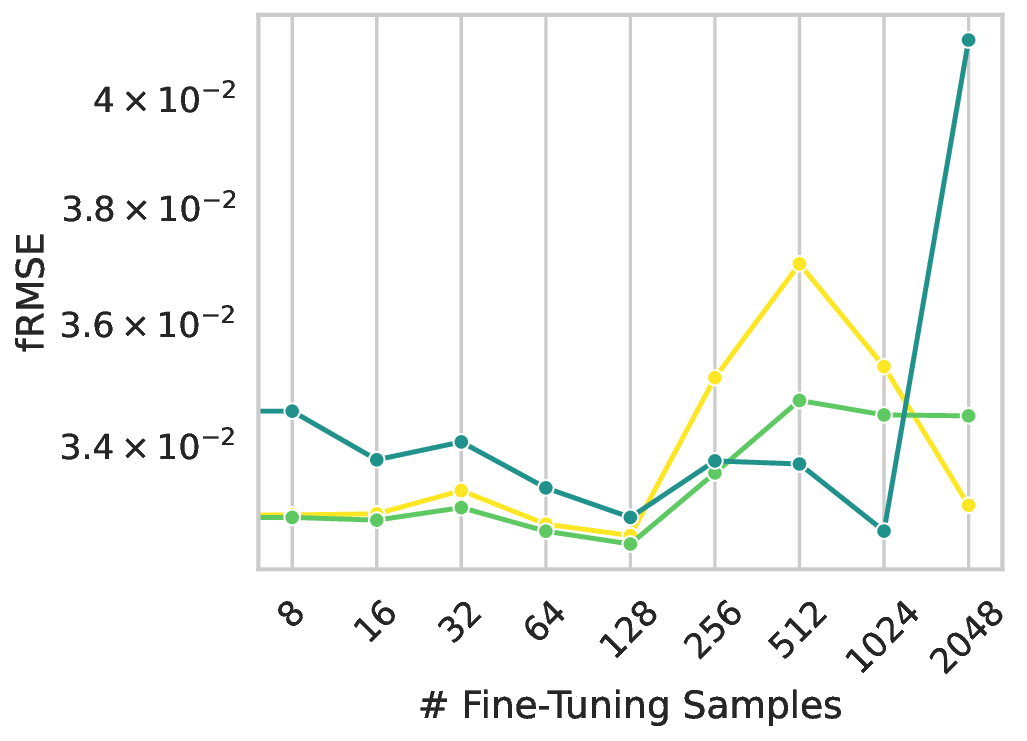}
    \caption{Medium}
    \label{fig:medium}
  \end{subfigure}\hfill
  \begin{subfigure}[b]{0.21\linewidth}
    \includegraphics[width=\linewidth]{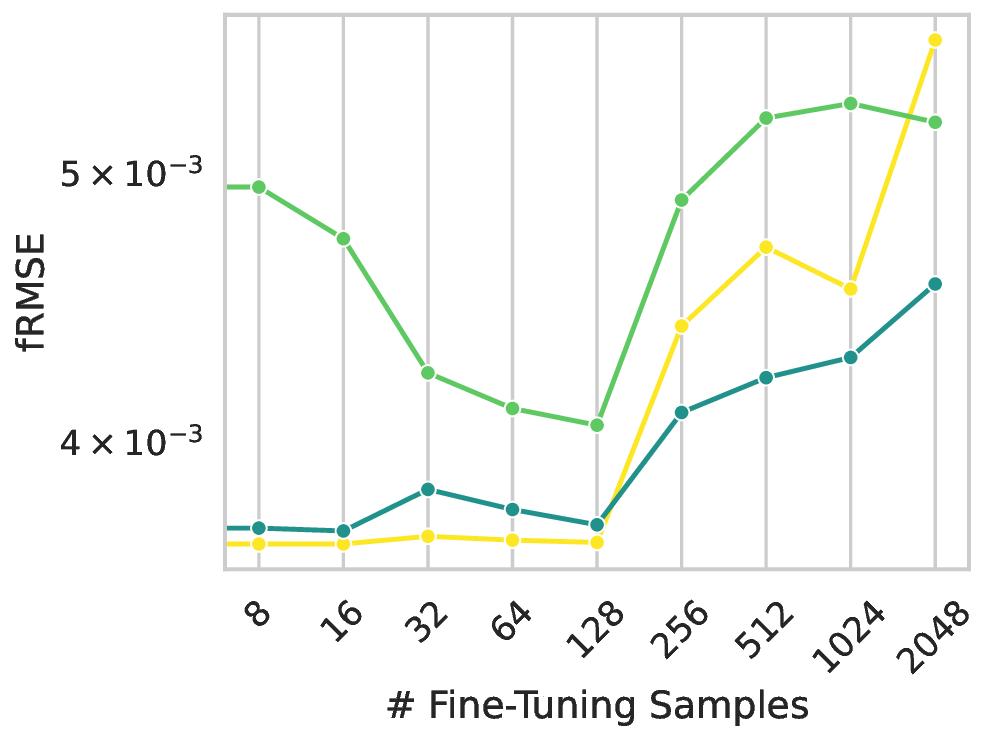}
    \caption{High}
    \label{fig:high}
  \end{subfigure}\hfill
  \begin{subfigure}[b]{0.32\linewidth}
    \includegraphics[width=\linewidth]{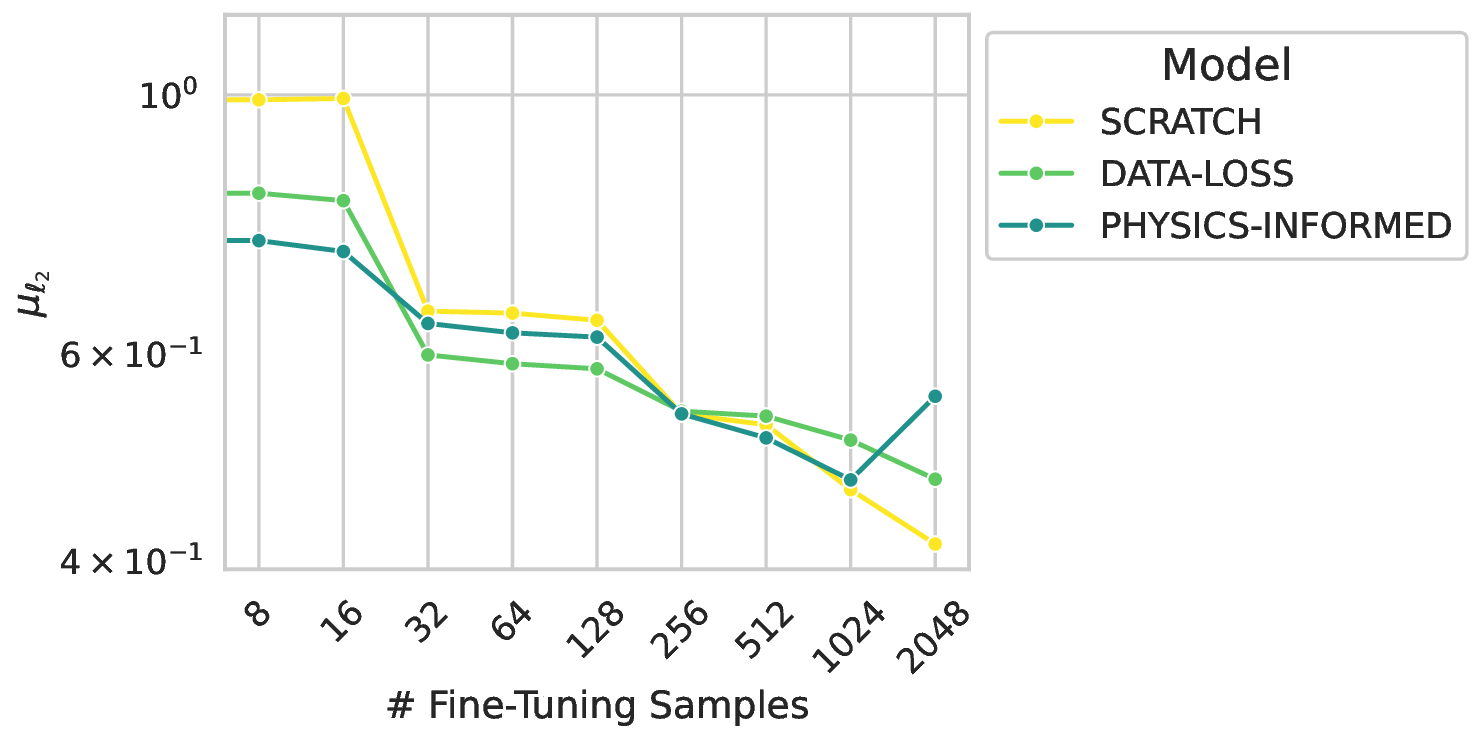}
    \caption{$\mu_{\ell_2}$}
    \label{fig:mul2}
  \end{subfigure}
  \caption{The fRMSE for low frequencies (a), medium frequencies (b), high frequencies (c), and $\mu_{\ell_2}$ loss (d) for the downstream task of \textbf{turbulent} Navier--Stokes.}
  \label{fig:new-operators-turbo-ns}
\end{figure}
We evaluate generalization to entirely new PDE classes not seen during pre-training: Reaction-Diffusion, Reaction-Advection-Diffusion, Navier-Stokes, and Darcy. These systems differ substantially in dynamics, operator structure, and boundary
conditions, shifting from steady-state to time-dependent problems, introducing new physical terms, and, for Darcy, replacing periodic with Dirichlet boundaries. They therefore constitute a strong test of whether the pre-trained representations are sufficiently general to support downstream adaptation.
The results, shown in Figure~\ref{fig:new-operators-extended-dataset-mul2}, reveal several key trends. For Reaction-Diffusion, the physics-informed pre-training improves over the data-only baseline. This is likely due to its similarity to pre-trained PDEs, particularly in its diffusive character. In contrast, for the more complex Reaction-Advection-Diffusion and Navier-Stokes equations, both pre-training strategies perform similarly, with neither offering a clear advantage. In the case of Darcy flow, with a distinct elliptic structure and different boundary conditions, the data-driven model outperforms the physics-informed model. These observations indicate that when the downstream PDE is substantially different from the pre-training distribution, the benefits of physics-informed pre-training diminish. 
Across all four unseen PDEs, both pre-trained models require substantial fine-tuning to match the performance of scratch-trained models, without consistently surpassing them. To further explore this limitation, we evaluate a more challenging case: \textbf{turbulent} Navier-Stokes, with fine-tuning limited to 2048 examples due to the high cost of data generation. As shown in Figure~\ref{fig:new-operators-turbo-ns} (d), both pre-trained models perform similarly to the scratch baseline, with the physics-informed model often underperforming. fRMSE analysis (Figure~\ref{fig:new-operators-turbo-ns} a–c) shows that all models struggle with low-frequency components, suggesting overfitting to dominant frequencies from pre-training. While performance improves with more data, this comes at the cost of reduced accuracy on mid- and high-frequency modes, leading to overshooting errors in the physics-informed model. 
Overall, these results indicate that current pre-training strategies lack the generalization needed for complex, high-frequency PDEs under limited data, highlighting the need for ``neural digital twins'' such as the NeurDE model~\cite{benitez2025neural} that in incorporate physical biases in a more meaningful way.
\section{Conclusion and Key Takeaways}

This study evaluated whether physics-informed pre-training, i.e., incorporating PDE residuals into the training loss, can improve the generalization, data efficiency, and robustness of models that aim to be SciFMs. To answer this, we ran systematic experiments across four settings: 
(i) adapting to new physics via parameter shifts,
(ii) generalizing to unseen PDE operators, 
(iii) transferring from simple to complex problems, and 
(iv) fine-tuning with noisy data. 
Our benchmark spans eight problems drawn from seven PDE families---Poisson, Advection-Diffusion, Helmholtz, Reaction-Diffusion, Reaction-Advection-Diffusion, Navier-Stokes (laminar and turbulent regimes), and Darcy
flow.

\medskip
\noindent
\textbf{Key takeaways:}
\begin{itemize}
    \item \textbf{In-scope gains.} Physics-informed pre-training improves generalization and data efficiency when downstream tasks strongly resemble the pre-training data, and it adds a measure of robustness under noisy fine-tuning.
    \item \textbf{Limited out-of-scope transfer.} The effectiveness deteriorates substantially for unseen PDE operators, non-periodic boundary conditions, and even moderately different dynamics, where physics-informed models can only match or trail data-only baselines.
    \item \textbf{Spectral overfitting.} Models tend to overfit to the dominant frequencies of the pre-training data, degrading on low- and high-frequency components which characterizes new settings.
    \item \textbf{Soft constraints are not enough.} Residual-based supervision alone does not instill transferable physical understanding; instead it only regularizes what the model learns. 
\end{itemize}
Overall, while physics-informed pre-training holds some promise, our results demonstrate that residual-based supervision is not sufficient on its own for building broadly transferable SciFMs. 
Building generalizable SciFMs requires stronger, more sophisticated, architectural integration of physical principles; the scale-coupling kinetic lifting approach underlying NeurDE~\cite{benitez2025neural} as well as recent results on Heavy-Tailed Self Regularization (HT-SR) Theory~\cite{MM19_HTSR_ICML,MM20_SDM,MM20a_trends_NatComm} suggest that a more principled approach will likely need to involve machine learning methods that incorporate scale-coupling inductive biases.

\begin{credits}
\subsubsection{\ackname} 
This work is funded by Deutsche Forschungsgemeinschaft (DFG, German Research Foundation) under Germany's Excellence Strategy - EXC 2075 – 390740016. We acknowledge the support by the Stuttgart Center for Simulation Science (SimTech). Additionally, the authors gratefully acknowledge the computing time provided on the high-performance computer HoreKa by the National High-Performance Computing Center at KIT (NHR@KIT). This center is jointly supported by the Federal Ministry of Education and Research and the Ministry of Science, Research and the Arts of Baden-Württemberg, as part of the National High-Performance Computing (NHR) joint funding program (https://www.nhr-verein.de/en/our-partners). HoreKa is partly funded by the German Research Foundation (DFG).

\subsubsection{\discintname}
The authors have no competing interests to declare that are relevant to the content of this article.

\end{credits}

\bibliographystyle{splncs04}
\bibliography{discoveryscience2026}

\appendix

\crefalias{section}{appendix}
\crefalias{subsection}{subappendix}
\crefalias{subsubsection}{subsubappendix}

\renewcommand{\theHsection}{\Alph{section}}
\renewcommand{\theHsubsection}{\theHsection.\arabic{subsection}}
\renewcommand{\theHsubsubsection}{\theHsubsection.\arabic{subsubsection}}
\clearpage

\section{Details of the Training and Evaluation PDE Systems}
\label{app:pde_details}
This appendix details the eight fundamental PDE systems summarized in \cref{pdes}, which serve as essential components of various scientific and engineering applications. For pre-training, we use three steady-state PDEs: the Poisson, Advection-Diffusion, and Helmholtz equations. For evaluation, we expand this set by five entirely new PDEs: the Reaction-Diffusion, Reaction-Advection-Diffusion, and Darcy equations, and the Navier-Stokes equations (both laminar and turbulent). Each system is defined over the spatial domain \( \Omega = [0,1]^2 \), while the time-dependent problems evolve over the interval \( \tau = [0,1] \). Except for the Darcy equation, which employs Dirichlet boundary conditions, all other PDEs assume periodic boundary conditions. The source term is represented as \( f(\mathbf{x}, t) \). The steady-state problems are designed to map the force function and PDE parameters to the corresponding steady-state solution. In contrast, the time-dependent problems aim to map the PDE parameters and initial conditions to the solution at \( t=1 \), as outlined below:\\[2mm]
\textbf{Poisson Equation:}
We consider a classical elliptic PDE that models diffusion processes with a source term:
\begin{equation}
    -\nabla \cdot \mathbf{D} \nabla u = f \quad \text{in } \Omega,
\end{equation}
where \( u(\mathbf{x}) \) is the unknown solution, and \( \mathbf{D} \) is the diffusion coefficient tensor that characterizes the material properties.\\ [2mm]
\textbf{Advection-Diffusion Equation:}
This equation describes the steady-state interplay between advective and diffusive transport processes:
\begin{equation}
    -\nabla \cdot \mathbf{D} \nabla u + \mathbf{v} \cdot \nabla u = f \quad \text{in } \Omega,
\end{equation}
where \( \mathbf{v} \) is the velocity field capturing advection effects, and \( \mathbf{D} \) is the diffusion coefficient tensor. To assess the balance between advection and diffusion, a dimensionless number is defined~\cite{shashank2023towards}:
\begin{equation}
    \Psi = \frac{\|\mathbf{v} \cdot \nabla u\|}{\|\nabla \cdot \mathbf{D} \nabla u\|}.
\end{equation}
\textbf{Helmholtz Equation:}
The Helmholtz equation models wave propagation and oscillatory phenomena commonly encountered in acoustics and electromagnetics:
\begin{equation}
    -\Delta u + \omega u = f \quad \text{in } \Omega,
\end{equation}
where \( \omega > 0 \) is the wave number that controls the frequency of oscillations in the solution.\\[2mm]
\textbf{Reaction-Diffusion Equation:}
This equation describes the distribution of chemical or biological species under diffusion and reaction processes:
\begin{equation}
    \frac{\partial u}{\partial t} - \nabla \cdot \mathbf{D} \nabla u + R(u) = f, \quad \text{in } \Omega \times \tau,
\end{equation}
where \( \mathbf{D} \) is the diffusion coefficient tensor, and the reaction term \( R(u) \) models localized changes due to interactions such as chemical reactions or biological processes.\\ [2mm]
\textbf{Reaction-Advection-Diffusion Equation:}
This PDE captures the combined effects of advection, diffusion, and reaction processes, which are commonly encountered in environmental and industrial applications:
\begin{equation}
    \frac{\partial u}{\partial t} -\nabla \cdot \mathbf{D} \nabla u + \mathbf{v} \cdot \nabla u + R(u) = f \quad \text{in } \Omega \times \tau,
\end{equation}
where \( \mathbf{v} \) is the velocity field capturing advection effects, \( \mathbf{D} \) is the diffusion coefficient tensor, and \( R(u) \) is the reaction term. \\ [2mm]
\textbf{Darcy Equation:}
The Darcy equation describes fluid flow through porous media and is fundamental in hydrogeology and petroleum engineering:
\begin{equation}
    \begin{split}
        \frac{\partial p}{\partial t} - \nabla \cdot (\mathbf{K} \nabla p) &= f \quad \text{in } \Omega \times \tau, \\
        \mathcal{B}(p) &= 0 \quad \text{in } \quad \mathbf{x} \in \partial\Omega \times \tau,
    \end{split}
\end{equation}
where \( p(\mathbf{x}) \) is the pressure field, and \( \mathbf{K} \) is the permeability tensor characterizing the medium's resistance to flow. For this problem setting, we consider the Dirichlet boundary condition. \\ [2mm]
\textbf{Navier-Stokes Equation:}
The Navier-Stokes equations are fundamental in fluid dynamics, providing insights into fluid behavior in various applications. We consider the viscous, incompressible fluid in vorticity form, which captures the dynamics of the fluid's vorticity and velocity fields:
\begin{equation}
    \begin{split}
        \frac{\partial u}{\partial t} + v \cdot \nabla u & = \nu \Delta u + f \quad \text{in } \Omega \times \tau, \\
        \nabla \cdot v  & = 0 \quad \text{in } \Omega \times \tau, \\
        u(\mathbf{x}, 0) & = u_0(\mathbf{x}) \quad \text{on } \Omega \\
    \end{split}
\end{equation}
where \( u \) represents the unknown vorticity field of the fluid, \( v \) is the velocity field, \( \nu \) is the kinematic viscosity coefficient that characterizes the fluid's resistance to flow, and \( f \) is the external force acting on the fluid. The continuity equation \( \nabla \cdot v = 0 \) ensures the incompressibility of the fluid. The problem is solved with periodic boundary conditions on \( \partial\Omega \). We consider two cases: (i) the force term \( f \) is zeroed and the initial condition \( u_0 \) is sampled as in the previous cases (laminar flow), and (ii) \( f \) is sampled as in the previous cases and \( u_0 \) is zeroed (turbulent flow).

\clearpage
\section{Implementation Details}
\label{app:implementation_details}
\subsection{Training and Inference Datasets Creation}
\label{subsec:datasets}

\paragraph{Discretization.}
While the inputs and outputs in general represent functions, such as the force function \( f \) and solution \( u \), we discretize them to allow us to work with them numerically. As our PDEs have two spatial dimensions, and for time-dependent problems one fixed value, we discretize these using an equidistant grid of points, with $h$ horizontal points and \( w\) vertical points. Thus, the functions are of the form
\begin{equation}
    f \in \mathbb{R}^{h \times w}, \quad \lambda \in \mathbb{R}^{h \times w \times c} \quad \text{and} \quad u \in \mathbb{R}^{h \times w},
\end{equation}
where \(c\) denotes the number of coefficient channels. For our experiments, we use a resolution of \(h \times w = 128 \times 128\).

\paragraph{Source Function Sampling.}
The source functions are generated using a linear combination of $n$ radial basis functions \( \{ \varphi_i(\mathbf{x}) \}_{i=1}^n\). Each basis function, \( \varphi_i(\mathbf{x}) \), is a Gaussian function centered at a point \( \mathbf{x}_i \in \Omega \) and defined as \(\varphi_i(\mathbf{x}) = \varphi(\mathbf{x} -\mathbf{x}_i)\). The source function is then defined as: \(f(\mathbf{x}) = \sum_{i=1}^{n} \varphi_i(\mathbf{x})p_i\), where \( p_i \) is sampled from a uniform distribution, \(p_i \sim \mathcal{U}(0, 1)\). The source function is further parameterized by a sparsity level \( s \), indicating the percentage of non-zero components, acting as a form of dropout. During all experiments, we fix the number of radial basis functions to \( n = 12 \times 12 = 144 \), each with a standard deviation of \( 1/32 \), while varying the sparsity level \( s \) between 20\%, 40\%, 60\%, and 80\%. This approach allows for generating a diverse set of force functions.

\paragraph{Coefficient Sampling.}
For all system coefficients, we target a description such that we are able to sample from a uniform distribution \( \mathcal U\). Coefficients, which are scalar quantities, such as the wave number \( \omega \in \mathbb N^+\), reaction term \( r \in \mathbb R\) and viscosity \( \nu \in \mathbb R\) can be directly sampled from a uniform distribution \( \mathcal U(\min, \max)\), where the tuple \((\min, \max)\) describes our target range. The two-dimensional velocity vector \(\mathbf{v} \in \mathbb R^2\) is sampled from \( \mathcal{U}(v_{\text{min}}, v_{\text{max}}) \) for magnitude and adjusted by a random angle \( \theta \sim \mathcal{U}(0, 2\pi) \) for direction. The diffusion coefficient tensor \(\mathbf{D} \in \mathbb{R}^{2\times 2}\) is constructed by sampling eigenvalues from a uniform distribution \( e \sim \mathcal{U}(e_{\text{min}}, e_{\text{max}}) \), leading to \(\mathbf{D} = R^T \text{diag}(e) R\), where the rotation matrix \( R = \text{rot}(\theta) \) is generated using a random angle \( \theta \sim \mathcal{U}(0, 2\pi) \). This allows control over the diffusion direction, while \(\text{diag}(e)\) governs anisotropy and diffusion extent. Finally, for the Advection-Diffusion equation, we sample from a fixed distribution range for the diffusion \( \mathbf{D} \sim (1,5)\) and velocity vector \( \mathbf{v} \sim (1,1)\). Additionally, we sample the dimensionless number \( \Psi \sim \mathcal{U}(\Psi_{\text{min}}, \Psi_{\text{max}}) \), which is then used to adjust the ratio between diffusion and velocity to achieve the desired balance.

\paragraph{Solution Generation.}
For the steady-state problems (Poisson, Advection-Diffusion, Helmholtz), given a sampled \( \lambda_k \) and \( f_k \), the solution \( u_k \) is computed using the spectral method. For the initial-value problems (Reaction-Diffusion, Reaction-Advection-Diffusion, Navier-Stokes), solutions are computed at \( t=1 \). We adapt the method used by~\cite{li2020fourier} for solving the Navier-Stokes equation, which combines the spectral method with a Crank--Nicolson update over time.

\paragraph{Dataset Composition.}
Depending on the use-case, the components of a dataset might change. Under normal circumstances, the dataset is composed of $N$ samples with \(\{(f_k, \lambda_k, u_k ) \mid f_k \in \mathbb{R}^{h \times w} ,  \lambda_k \in \mathbb{R}^{h \times w \times c}, u_k \in \mathbb{R}^{h \times w} \}_{k=1}^{N}\). For time-dependent problems, each sample additionally carries the initial condition \( u^0_k \in \mathbb{R}^{h \times w} \), matching the fine-tuning dataset \( \mathcal{D}_{\text{fine}} \) introduced in \cref{sec:method}.
For a given sample and PDE operator \( \mathcal{G}_i \), \( f_k \) represents the force function for this sample, \( \lambda_k \) represents the coefficients of the underlying PDE and \( u_k \) the corresponding ground-truth.
PDEs might be parameterized by multiple coefficients, with varying dimensionalities. Due to this, we decompose all coefficients into scalar quantities and assign them separate channels \(c\), with the constant function of the scalar value. For example, the diffusion coefficient \( \mathbf{D} \in \mathbb{R}^{2 \times 2} \) is decomposed into four scalar values (\( \mathbf{D}_{(1, 1)}, \mathbf{D}_{(1, 2)}, \mathbf{D}_{(2, 1)}, \mathbf{D}_{(2, 2)}\)). As a result, the coefficients form 4 separate channels, yielding a tensor of shape \(\mathbb{R}^{h \times w \times 4}\), where for all points in channel 1 the value is \(\mathbf{D}_{(1, 1)}\), for channel 2 the value is \(\mathbf{D}_{(1, 2)}\), and so on. The channel size is denoted by \( c \).
Further, a dataset might be constructed from multiple datasets with different PDEs. In these cases, the coefficient component \( \lambda \) is extended to contain the same number of channels as the number of different coefficients needed for all PDEs (e.g., a dataset comprising Poisson and Advection-Diffusion will have a coefficient channel size of 6 due to the diffusion tensor \( \mathbf{D} \in \mathbb{R}^{2 \times 2} \) found in both PDEs and the additional \( \mathbf{v} \in \mathbb{R}^{2}\) for the velocity term in the Advection-Diffusion case). Coefficient channels which are not needed for a sample to fully represent the PDE are zeroed.

\paragraph{Pre-training Datasets.}
The pre-training datasets are specifically designed to train the foundation models on a broader range of partial differential equations, enabling the models to learn generalizable features and patterns, which include data for the Poisson, Advection-Diffusion, and Helmholtz equations only. This phase is crucial for establishing a robust initial model capable of effectively capturing the underlying dynamics of various PDEs. The pre-training datasets are constructed from the PDEs outlined in \cref{tab:pdes}.

\paragraph{Fine-Tuning Set.}
For the downstream tasks, we consider only single PDE systems as targets, including the pre-training targets Poisson, Advection-Diffusion, and Helmholtz and additionally the Reaction-Diffusion, Reaction-Advection-Diffusion, Navier-Stokes and Darcy tasks. For tasks included in the pre-training, we consider different coefficient ranges, i.e., ID, Slight-OOD, Medium-OOD, and High-OOD, shifted with regards to the pre-training set. This allows us to evaluate the models' ability to generalize within the same operator, either under similar or significantly different dynamics. PDEs not included in the pre-training set allow us to evaluate how well the models generalize towards new operators. For the Darcy target specifically, we use the datasets provided by PDEBench~\cite{takamoto2022pdebench}\footnote{Available at \url{https://doi.org/10.18419/darus-2986}.}, combining all to create a diverse downstream target.

\paragraph{Dataset Partitioning.}
For each target PDE, we create a training set of \( 2^{15} \) samples, and validation and test sets of \( 2^{12} \) samples each, all at \( 128 \times 128 \) resolution. Since the pre-training set consists of three PDEs, the total pre-training sample counts are \( 3 \times 2^{15} \) (training) and \( 3 \times 2^{12} \) (validation/test). When a training regime requires fewer samples, we subsample the full dataset.

\paragraph{Input Normalization.}
We incorporate input normalization, as proposed in~\cite{shashank2023towards}, as it enhances FNO model performance, particularly when input channels have different scales. For this, we first compute a reference force norm \( f_{\text{ref}} := \text{median}_k(|| f_k ||_2) \) over the corresponding dataset. Then, for each sample \((f_k, \lambda_k)\) in the dataset, we normalize the force function and coefficient component as follows:

\begin{equation}
    \tilde{f}_k = f_k \frac{f_{\text{ref}}}{|| f_k ||_2} \quad \text{and} \quad \tilde{\lambda}_k = \lambda_k \frac{f_{\text{ref}}}{|| f_k ||_2}.
    \label{eq:input-normalization}
\end{equation}

This ensures that the norms of the force functions are consistent across the dataset, bringing the input features onto a similar scale. This normalization helps the model learn an operator whose output is less sensitive to the original scale of the input functions.

\paragraph{Noise.}
To evaluate the robustness of the model against noise during the fine-tuning process, we introduce noise into the solution space of the fine-tuning dataset, simulating real-world measurement inaccuracies. Specifically, we introduce Additive White Gaussian Noise into each ground-truth solution \( u \) in the dataset, formulated as:

\begin{equation}
    u_{\text{noisy}} = u + \sigma \cdot \text{std}(u) \cdot \epsilon,
    \label{eq:awgn}
\end{equation}

where \( \epsilon \sim \mathcal{N}(0, 1) \) represents a standard normal random variable, and \( \sigma \) denotes the noise level. The term \( \text{std}(u) \) is the standard deviation of the ground-truth solution \( u \). Throughout the experiments, we use four noise levels, fixing the level per dataset:  \( \sigma = 0.01 \), simulating high-precision measurements, \( \sigma = 0.05 \) and \( \sigma = 0.1 \) to reflect typical measurement uncertainties, and \( \sigma = 0.2 \) to represent highly noisy conditions. By simulating real-world measurement inaccuracies, this technique allows us to evaluate how well the model can maintain its performance in the presence of inaccuracies, thereby providing insights into its reliability in practical applications.
\subsection{Model Architecture Selection}
We evaluate two model families for physics-informed pre-training: the Fourier Neural Operator (FNO)~\cite{li2020fourier} and the Axial Vision Transformer (AViT)~\cite{mccabe2024multiple}. Both models are trained using our physics-informed framework and tested in a zero-shot setting across multiple downstream PDE tasks. In addition to the \( \mu_{\ell_2} \) error introduced in \cref{subsec:evaluation_metrics}, we report the \( L_{\infty} \) error, which measures the maximum deviation between the predicted and ground truth solutions, defined as \( L_{\infty} = \max_{i} \mid u_i - \hat{u}_i \mid \). FNO consistently outperforms AViT in all scenarios (see \cref{tab:avit_vs_fno}). This performance gap may be attributed to FNO's spectral architecture, which provides more accurate and stable derivative computations, crucial for evaluating the PDE residuals used in the physics-informed loss. Due to its superior zero-shot accuracy and computational efficiency, we use the FNO architecture in most subsequent experiments.
\begin{table*}[!htbp]
    \caption{Comparison of FNO and AViT architectures (approximately 67M parameters) evaluated on coefficient ranges progressively pushed OOD from the pre-training distribution. Bold entries indicate better performance. Predictions with error $>$ 0.2 are considered failures and omitted from analysis.}
    \label{tab:avit_vs_fno}
    \resizebox{\textwidth}{!}{%
        \begin{tabular}{@{}lllllllllllllll@{}}
        \toprule
        \multirow{2}{*}{Metric}         & \multirow{2}{*}{Architecture} & \multirow{2}{*}{Model} & \multicolumn{4}{l}{Poisson}                                              & \multicolumn{4}{l}{Advection-Diffusion}                      & \multicolumn{4}{l}{Helmholtz}                               \\ \cmidrule(l){4-15}
        &                               &                        & ID              & Slight-OOD       & Medium-OOD       & High-OOD         & ID               & Slight-OOD       & Medium-OOD       & High-OOD         & ID               & Slight-OOD       & Medium-OOD & High-OOD \\ \midrule
        $\mu_{\ell_2}$ & FNO      & Physics-informed            & \textbf{0.0005} & \textbf{0.0067} & \textbf{0.1326}           & 0.7091           & \textbf{0.0025} & \textbf{0.0260} & 1.2705           & 4.1350           & \textbf{0.0007}  & \textbf{0.2363}  & 1.2177     & 1.1825   \\ \cmidrule(l){2-15}
        & AViT         & Physics-informed            & 0.2852          & 0.3476           & 0.4591           & 0.5693 & 0.3166           & 0.3899           & 0.5949 & 0.7527 & 0.7792           & 0.9276           & 0.9834     & 0.9809   \\ \midrule
        $L_{\infty}$   & FNO          & Physics-informed            & \textbf{0.0007} & \textbf{0.0124} & \textbf{0.0836}  & 0.3155           & \textbf{0.0023} & \textbf{0.1325}           & 1.6630           & 2.7702           & \textbf{0.0036}  & 1.5212           & 1.6534     & 0.6385   \\ \cmidrule(l){2-15}
        & AViT         & Physics-informed            & 0.2635          & 0.2098           & 0.2266           & 0.2487 & 0.2432           & 0.2487           & 0.6253 & 1.4406           & 6.2318           & 5.9284          & 1.5243     & 0.5796   \\ \bottomrule
        \end{tabular}%
    }
\end{table*}

\subsection{Models}
In our experimental setup, we implement three FNO models, where the model's embedding dimension and Fourier modes are fixed at 64 and 32, respectively. These values strike a favorable balance between performance and computational cost.
\begin{itemize}
    \item \textbf{Scratch.} This model serves as our first baseline. It is trained directly on each corresponding downstream task without any pre-training, allowing us to assess the impact of the pre-training phase itself.

    \item \textbf{Data-Loss.} This model serves as a data-driven pre-training baseline. It is pre-trained using only the data loss \(\mathcal{L}_{\text{data}}\), following that of~\cite{shashank2023towards}. Subsequently fine-tuned on each downstream task, this model helps us assess the impact of incorporating physics-based constraints during pre-training on generalization.

    \item \textbf{Physics-Informed.} This model combines data-driven and physics-based signals using the hybrid loss \(\mathcal{L}_{\text{hybrid}}\) during pre-training. We fix the weighting factor \(\alpha = 0.5\) for equal contribution of both signals.
\end{itemize}
All inputs to the models are two-dimensional tensors, discretized on a grid of size \(h \times w = 128 \times 128\), with the source function and coefficients serving as input channels $c$, forming an input tensor of shape $\mathbb{R}^{h \times w \times c}$.
The models' output represents the PDE's solution space as a two-dimensional tensor of shape $\mathbb{R}^{h \times w}$.
The pre-trained models have an input dimension of $c = 8$, corresponding to the number of source and coefficient channels required for all pre-training tasks. In contrast, the input channel dimension for the Scratch model is adjusted to reflect the minimum number of channels necessary for the downstream tasks.
\subsection{Training and Inference Procedure}
For all training stages, we use the Adam optimizer with a cosine learning rate decay, training for a total of 500 epochs. The model with the lowest validation loss is saved for subsequent evaluation. Further, we use identical hyperparameters for all models during each task.
During pre-training, the Data-Loss and Physics-Informed models are trained on the same tasks, a mixed set of Poisson, Advection-Diffusion, and Helmholtz equations, comprising a total of \( 2^{15} \times 3 \) samples. During this stage, we vary the loss function according to the model being trained. The Data-Loss model is trained solely with the \( \mathcal{L}_{\text{data}} \) data loss, while the Physics-Informed model is trained with both the physics loss term and the data loss term.
For downstream tasks, all models are specifically trained on the target PDE dataset using only the \( \mathcal{L}_{\text{data}} \) data loss. The dataset sizes are referred to as Downstream Examples, indicating the number of samples used for training in these tasks. The Scratch model is trained directly on the downstream tasks without any pre-training.

\noindent
In cases where the downstream tasks do not include coefficients from the pre-training stages, the corresponding coefficient input channels are set to zero. Additionally, if the downstream task introduces new coefficient terms, a zero-shot test is conducted to determine which channel to use for the new coefficient based on the lowest validation loss achieved. The models are validated and tested on the complete validation and test sets, respectively, using a setup identical to the training of the models.

\clearpage

\section{Complementary Results}
\label{app:complementary_results}
In this appendix, we present complementary analyses that clarify and extend the results of the main text. For completeness and coherence, we may occasionally revisit material already discussed.
\subsection{Zero-Shot Performance Analysis}
We compare the zero-shot generalization performance of the Data-Loss and Physics-Informed models on downstream tasks featuring the same PDE operators used during pre-training but with varying coefficient ranges. This analysis evaluates the efficacy of physics-informed pre-training for improving generalization. \cref{app_tab_zero_shot} summarizes the performance using the \(\mu_{\ell_2}\) and \( L_{\infty}\) metrics on the ID and Slight-, Medium-, and High-OOD coefficient ranges.
\begin{table*}[!htbp]
    \centering
    \caption{Zero-shot evaluation on downstream tasks governed by the pre-training PDEs, under in-distribution and slight, medium, and high out-of-distribution (S/M/H-OOD) parameterizations, using the \(\mu_{\ell_2}\) and \( L_{\infty}\) metrics. Best per metric in bold.}
    \resizebox{\linewidth}{!}{%
        \begin{tabular}{llllllllllllll}
        \toprule
        \multirow{2}{*}{Metric}           & \multirow{2}{*}{Model} & \multicolumn{4}{l}{Poisson}                                           & \multicolumn{4}{l}{Advection-Diffusion}                               & \multicolumn{4}{l}{Helmholtz}                                         \\
                                          &                        & ID            & Slight-OOD           & Medium-OOD           & High-OOD           & ID            & Slight-OOD           & Medium-OOD           & High-OOD           & ID            & Slight-OOD           & Medium-OOD           & High-OOD           \\ \midrule
        \multirow{3}{*}{$\mu_{\ell_2}$}   & Scratch                   & 1.0080          & 1.1736          & 1.0061          & 1.8586          & 1.0005          & 1.1850          & 1.2476    & \textbf{1.3294} & 1.0137          & 1.1039          & 1.5014          & 1.3249          \\
                                          & Data-Loss                   & 0.0025          & 0.0102          & \textbf{0.1225} & \textbf{0.5750} & 0.0068          & 0.0434          & \textbf{0.8813} & 1.8948    & 0.0040          & 0.2448    & \textbf{1.2165} & 1.2121    \\
                                          & Physics-Informed                  & \textbf{0.0005} & \textbf{0.0067}    & 0.1326    & 0.7091    & \textbf{0.0025}    & \textbf{0.0260}    & 1.2705          & 4.1350          & \textbf{0.0007} & \textbf{0.2363} & 1.2177    & \textbf{1.1825} \\ \midrule
        \multirow{3}{*}{\( L_{\infty} \)} & Scratch                    & 0.8171          & 0.4959          & 0.4243          & 0.4702          & 0.6644          & 0.6286          & \textbf{0.9239} & \textbf{1.6784} & 7.8970          & 7.5244          & \textbf{1.5851} & 0.6475    \\
                                          &  Data-Loss                  & 0.0035          & 0.0182          & 0.1661          & 0.7197    & 0.0060          & \textbf{0.1172}    & 1.0009    & 2.0415    & 0.0557          & \textbf{1.5207} & 1.6423    & 0.6822          \\
                                          & Physics-Informed & \textbf{0.0007} & \textbf{0.0124}    & \textbf{0.0836} & \textbf{0.3155} & \textbf{0.0023}    & 0.1325          & 1.6630          & 2.7702          & \textbf{0.0036} & 1.5212    & 1.6534          & \textbf{0.6385} \\ \bottomrule
        \end{tabular}%
    }
    \label{app_tab_zero_shot}
\end{table*}

\begin{figure}[!htbp]
    \centering
    \includegraphics[width=1\textwidth]{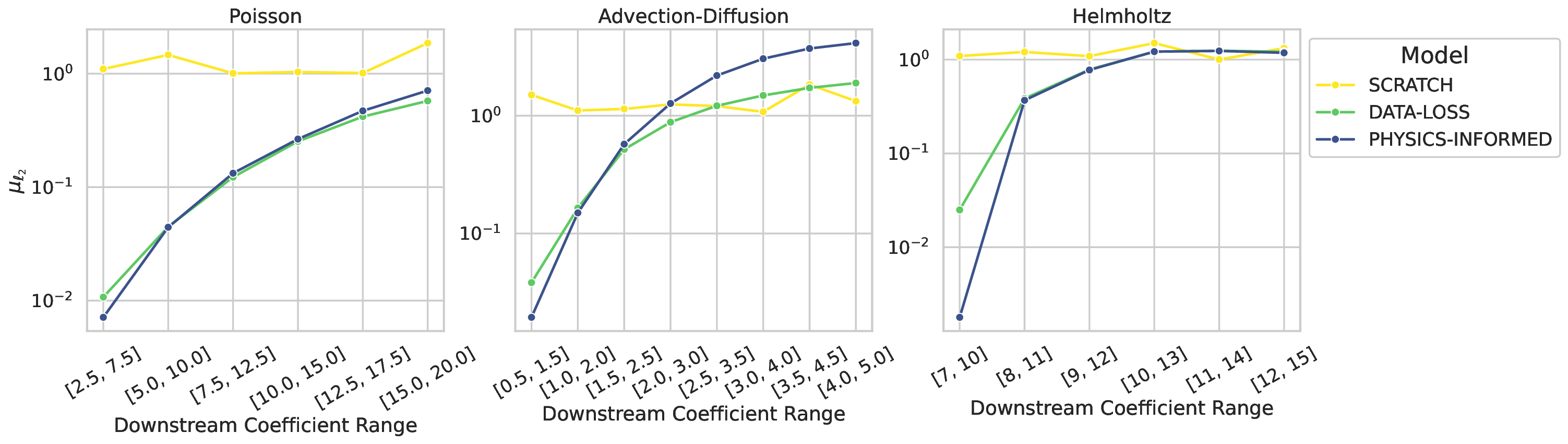}
    \caption{The $\mu_{\ell_2}$ metric for the downstream tasks of Poisson, Advection-Diffusion, and Helmholtz, respectively, where the coefficient ranges are gradually pushed OOD.}
    \label{appfig:zero-shot:expensive:mul2}
\end{figure}

\begin{figure}[!htbp]
    \centering
    \includegraphics[width=1\textwidth]{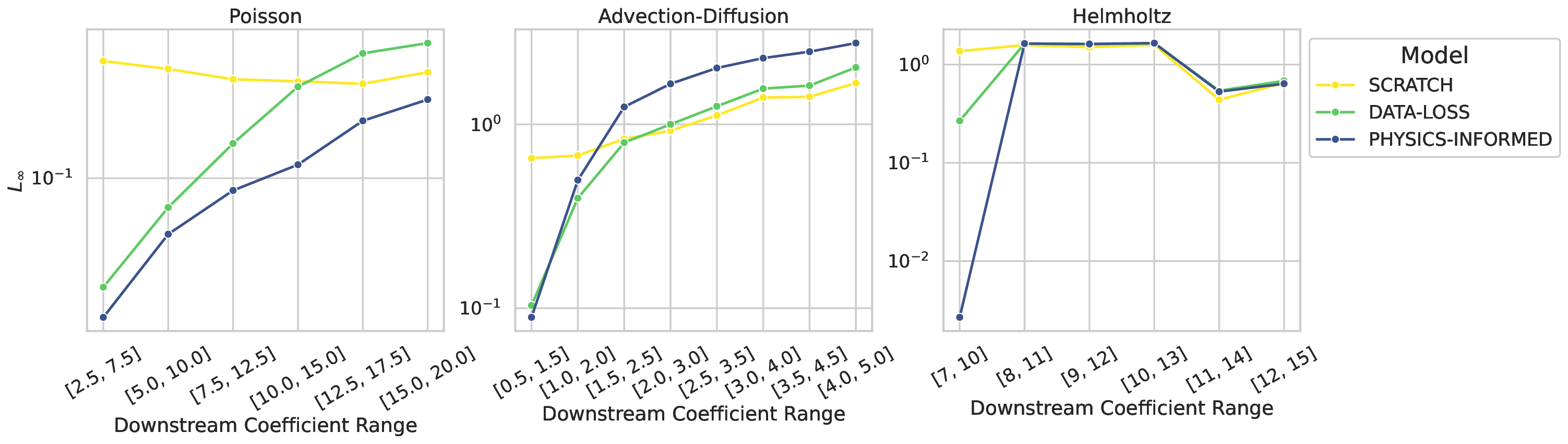}
    \caption{The $L_{\infty}$ metric for the downstream tasks of Poisson, Advection-Diffusion, and Helmholtz, respectively, where the coefficient ranges are gradually pushed OOD.}
    \label{appfig:zero-shot:expensive:linf}
\end{figure}

Both pre-training methods demonstrate significant improvements in zero-shot generalization, compared to the Scratch model, in both the ID and Slight-OOD settings. The results for the Data-Loss model are consistent with the findings of~\cite{shashank2023towards}. The Physics-Informed approach also demonstrates strong effectiveness in the Helmholtz ID and Slight-OOD settings compared to the Data-Loss model. This suggests that including PDE residuals as a training signal helps models better capture the underlying physical structure of the PDEs. In the Helmholtz task, slight variations in the wave number drastically alter the solution space, creating a more challenging OOD scenario. Similar effects can be observed in the Advection-Diffusion task, where significantly shifting the coefficient range out-of-distribution compared to the pre-training dataset introduces high velocities into the system, altering the solution space significantly compared to that of ID. Under these conditions, the Physics-Informed model's performance deteriorates, indicating that the introduced \( \mathcal L_\text{PDE} \) is sensitive to instances where the solution space diverges considerably from the pre-training data. Meanwhile, the Data-Loss model demonstrates greater robustness to variations in coefficient ranges.
%
We further evaluate the zero-shot performance of models by examining performance across progressively out-of-distribution coefficient ranges, providing a fine-grained picture of their failure modes. The results for the \( \mu_{\ell_2} \) and \( L_{\infty}\) metrics are shown in \cref{appfig:zero-shot:expensive:mul2,appfig:zero-shot:expensive:linf}, respectively.
The results illustrate how sensitive the Physics-Informed model is to shifts in the coefficient ranges which induce significant changes in the solution space. For the Poisson equation, the shift in coefficient increases the diffusion coefficient \( \mathbf{D} \), which does not change the solution space significantly compared to its ID counterpart. Yet, the model's \( \mu_{\ell_2} \) rises more quickly. This is even more exacerbated in the Advection-Diffusion and Helmholtz tasks.
For larger parameter shifts, the Physics-Informed model shows no improvement over the data-driven baseline. This indicates failure to enhance generalization in these regimes, suggesting the model does not effectively learn transferable physical principles beyond the pre-training distribution.
\subsection{n-Shot Learning}
\begin{figure}[!htbp]
    \centering
    \includegraphics[width=0.9\linewidth]{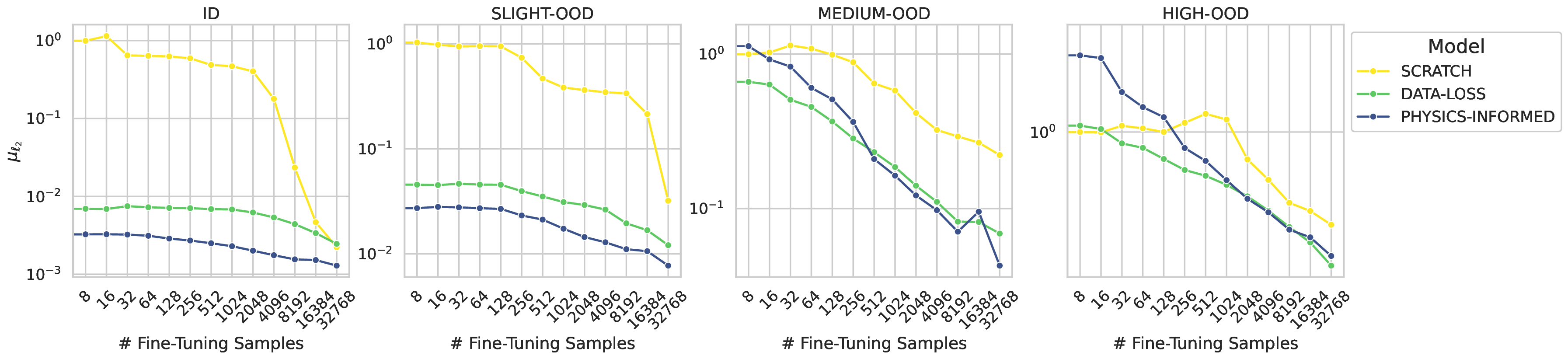}
    \caption{Evaluation of models fine-tuned with varying numbers of downstream samples for the n-shot Advection-Diffusion task, using the \(\mu_{\ell_2}\) metric. The coefficient ranges for the ID, Slight-, Medium-, and High-OOD settings follow \cref{tab:pdes}.}
    \label{appfig:n-shot:base:expensive:ad:mul2}
\end{figure}

\begin{figure}[!htbp]
    \centering
    \includegraphics[width=0.9\linewidth]{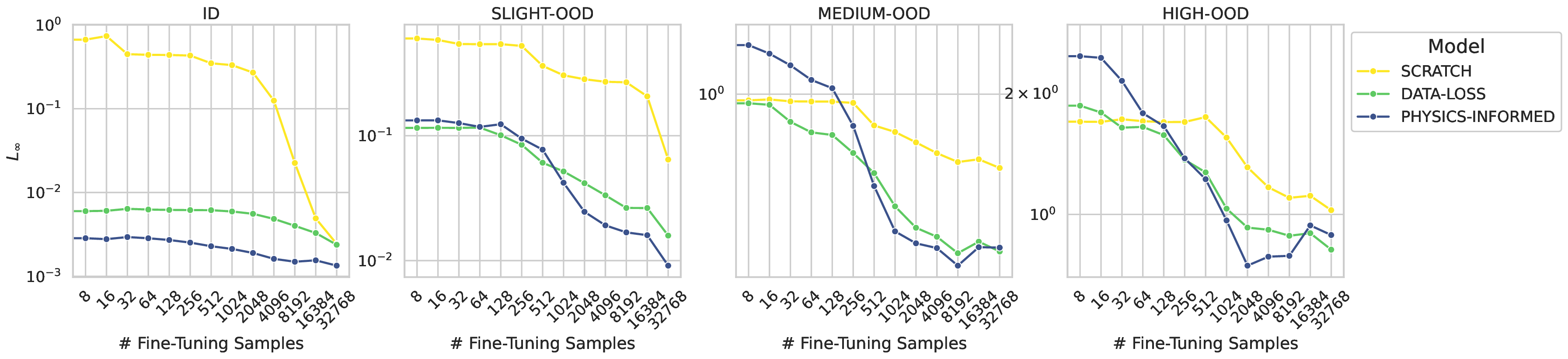}
    \caption{Evaluation of models fine-tuned with varying numbers of downstream samples for the n-shot Advection-Diffusion task, using the \( L_{\infty}\) metric.}
    \label{appfig:n-shot:base:expensive:ad:linf}
\end{figure}

\begin{figure}[!htbp]
    \centering
    \includegraphics[width=\linewidth]{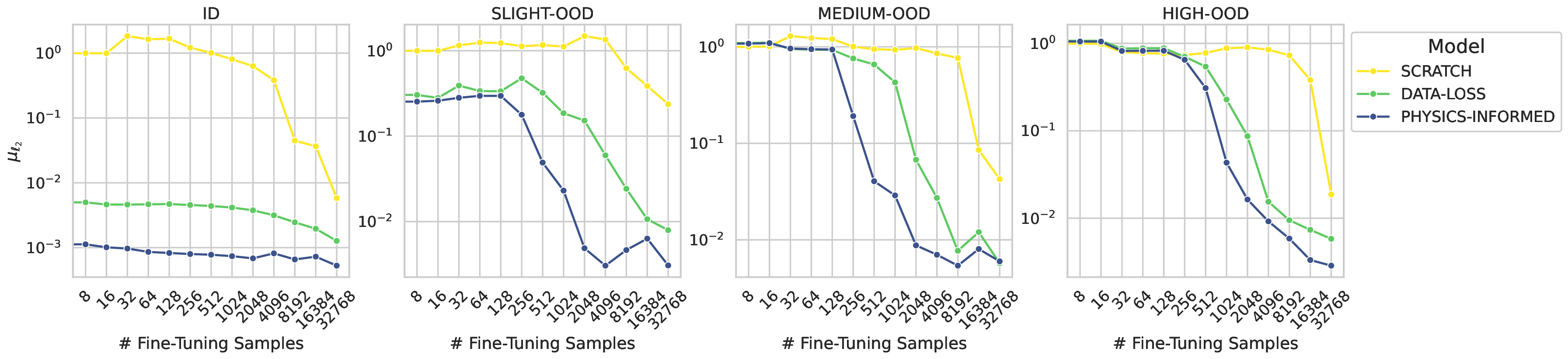}
    \caption{Evaluation of models fine-tuned with varying numbers of downstream samples for the n-shot Helmholtz task, using the \(\mu_{\ell_2}\) metric.}
    \label{appfig:n-shot:base:expensive:helm:mul2}
\end{figure}

\begin{figure}[!htbp]
    \centering
    \includegraphics[width=\linewidth]{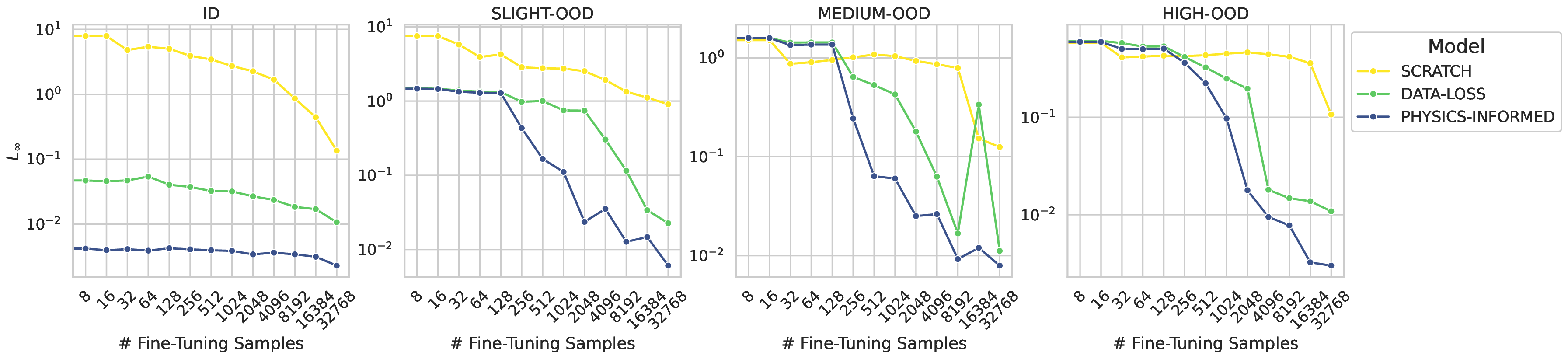}
    \caption{Evaluation of models fine-tuned with varying numbers of downstream samples for the n-shot Helmholtz task, using the \( L_{\infty}\) metric.}
    \label{appfig:n-shot:base:expensive:helm:linf}
\end{figure}

This subsection presents the n-shot evaluation of the Data-Loss baseline and the Physics-Informed model, focusing on their n-shot capabilities on the Advection-Diffusion and Helmholtz tasks. For the Advection-Diffusion equation, the \(\mu_{\ell_2}\) and \( L_{\infty}\) results are shown in \cref{appfig:n-shot:base:expensive:ad:mul2,appfig:n-shot:base:expensive:ad:linf}, respectively. Similarly, the \(\mu_{\ell_2}\) and \( L_{\infty}\) results for the Helmholtz equation are presented in \cref{appfig:n-shot:base:expensive:helm:mul2,appfig:n-shot:base:expensive:helm:linf}, respectively. Finally, we decompose the errors into frequency bands to substantiate the spectral-overfitting interpretation of \cref{subsec:n-shot-learning}.

\begin{figure}[!htbp]
    \centering
    \includegraphics[width=\linewidth]{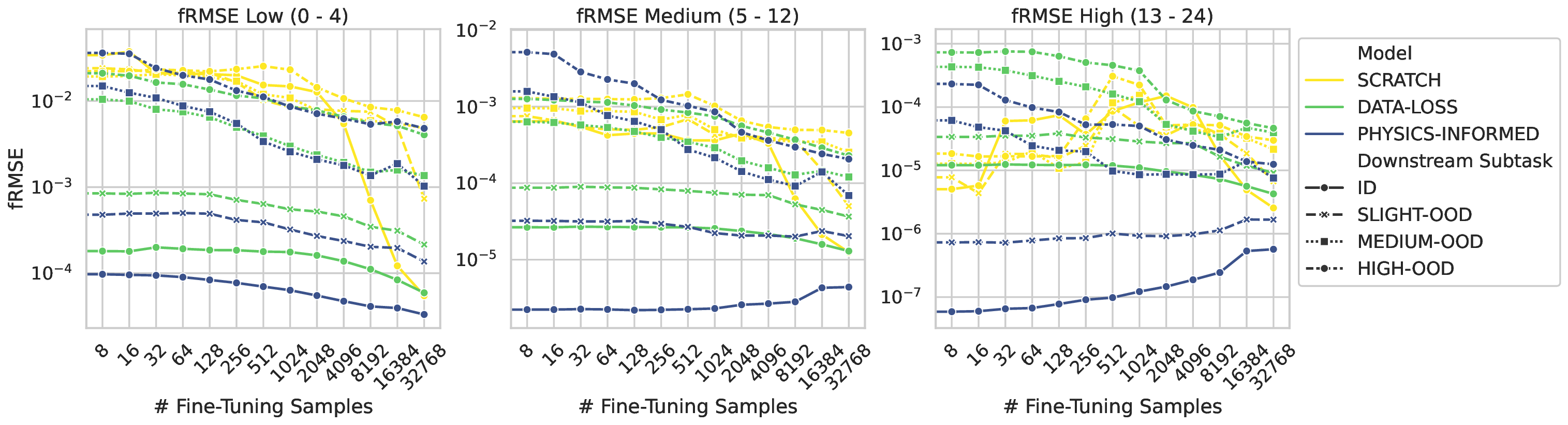}
    \caption{Evaluation of models fine-tuned with varying numbers of downstream samples for the n-shot Advection-Diffusion task, using the fRMSE metric for low-, medium-, and high-frequency bands.}
    \label{appfig:n-shot:base:expensive:ad:frmse}
\end{figure}

\begin{figure}[!htbp]
    \centering
    \includegraphics[width=\linewidth]{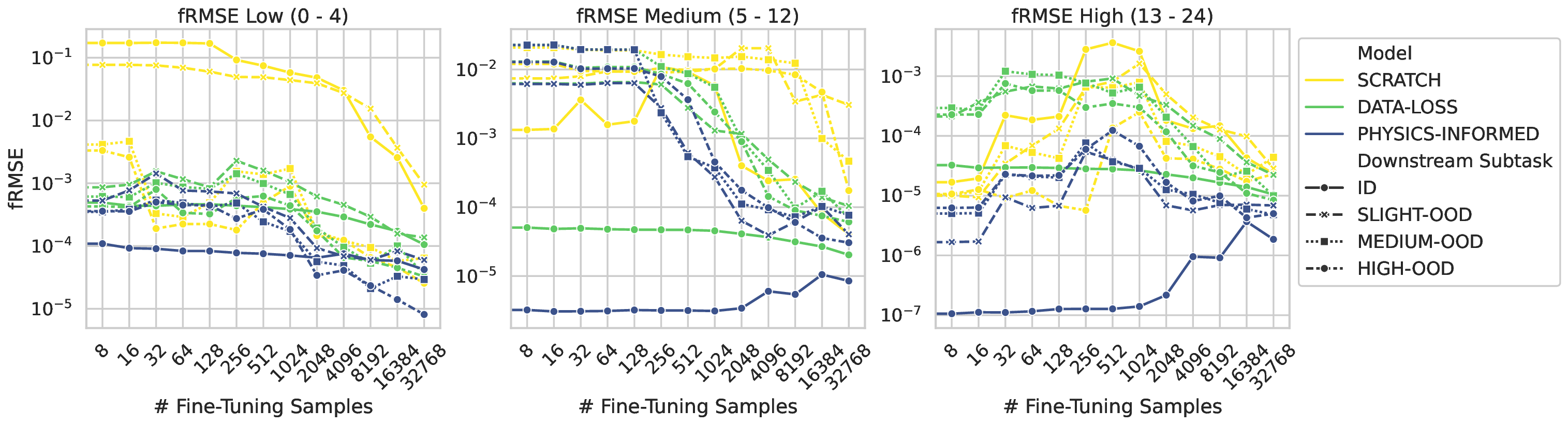}
    \caption{Evaluation of models fine-tuned with varying numbers of downstream samples for the n-shot Helmholtz task, using the fRMSE metric for low-, medium-, and high-frequency bands.}
    \label{appfig:n-shot:base:expensive:helm:frmse}
\end{figure}

\paragraph{Advection-Diffusion.}
In ID and Slight-OOD settings, Physics-Informed models outperform the Data-Loss baseline in low-data regimes, demonstrating improved sample efficiency consistent with zero-shot results (\cref{appfig:n-shot:base:expensive:ad:mul2,appfig:n-shot:base:expensive:ad:linf}). The benefit is particularly notable in the \(L_{\infty}\) error, which is critical in applications sensitive to worst-case deviations. However, in Medium- and High-OOD settings, this advantage diminishes. Physics-Informed models perform worse under low-data conditions, likely due to poor adaptability to the shifted solution space. While performance improves with more data, eventually surpassing the Data-Loss baseline at higher sample counts, the gains are delayed and less consistent. Across all regimes, pre-trained models still require fewer samples than training from scratch to reach comparable accuracy.
\paragraph{Helmholtz.}
For the Helmholtz equation, Physics-Informed models perform well in the ID setting (\cref{appfig:n-shot:base:expensive:helm:mul2,appfig:n-shot:base:expensive:helm:linf}). However, in low-data OOD regimes, their performance drops, indicating sensitivity to distribution shifts and overfitting to the pre-training setting. As the number of fine-tuning samples increases, the Physics-Informed model achieves notably better performance in the Medium-OOD case and moderately better performance in the High-OOD case. Compared to the Scratch model, both pre-training approaches require up to $8\times$ fewer samples to reach similar accuracy.
\paragraph{Spectral Overfitting.}
To substantiate the spectral-overfitting interpretation put forward in \cref{subsec:n-shot-learning}, we decompose the n-shot errors into low-, medium-, and high-frequency bands. The fRMSE results for the Advection-Diffusion and Helmholtz tasks are shown in \cref{appfig:n-shot:base:expensive:ad:frmse,appfig:n-shot:base:expensive:helm:frmse}, respectively. In the ID and Slight-OOD settings, the Physics-Informed model retains consistently lower fRMSE values than both baselines across all frequency bands; the gap is most pronounced for the high-frequency components, where its error remains up to two orders of magnitude below the Data-Loss baseline. However, these gains do not carry over to the Medium- and High-OOD settings: there, the Physics-Informed model performs comparable to, or worse than, the Data-Loss baseline in low-data regimes, and the differences between all models diminish only with substantial fine-tuning. Taken together, this frequency-resolved view suggests that physics-informed pre-training aligns the model with the dominant frequency content of the pre-training distribution, providing an advantage precisely when the downstream spectrum is familiar, and adapting poorly to new spectral patterns.
\subsection{n-Shot Noisy Learning}

This subsection extends the analysis of model performance in the n-shot learning setting by evaluating robustness to different degrees of noise in the solution data used for fine-tuning. We evaluate the models on the Advection-Diffusion and Helmholtz tasks, as they represent significantly more challenging settings compared to Poisson. For this analysis, we systematically increase the noise level from \( \sigma = 0.01 \) to \( \sigma = 0.2 \) and evaluate performance across various out-of-distribution settings. The in-distribution setting is excluded from this evaluation.

\begin{figure}[!htbp]
    \centering
    \includegraphics[width=\linewidth]{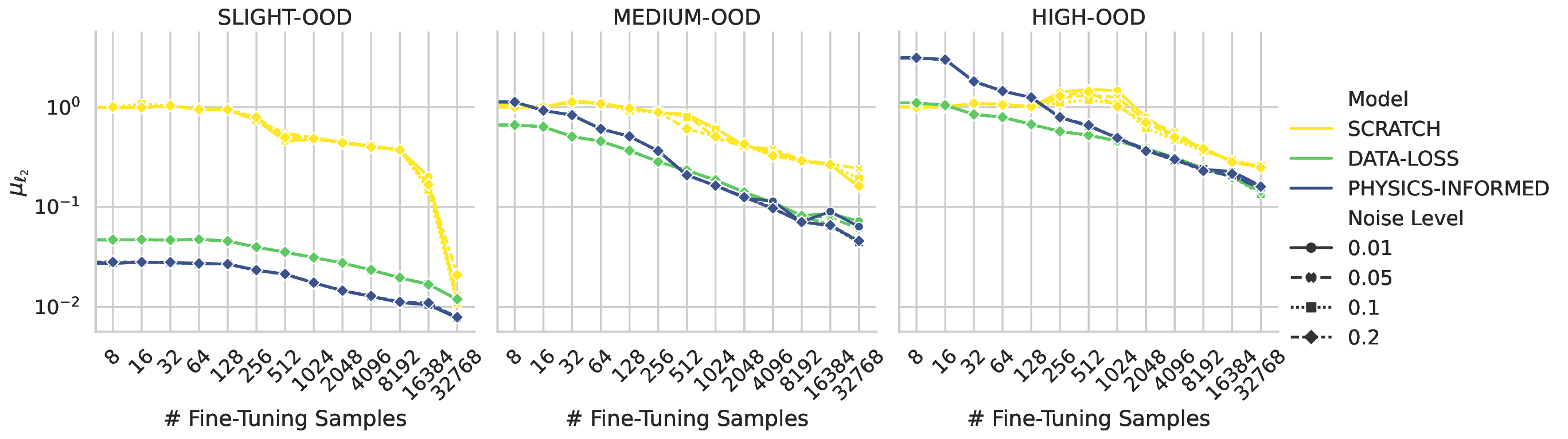}
    \caption{Evaluation of models fine-tuned with varying numbers of noisy downstream samples for the n-shot Advection-Diffusion task, using the \(\mu_{\ell_2}\) metric.}
    \label{appfig:n-shot:robustness:ad:mul2}
\end{figure}

\begin{figure}[!htbp]
    \centering
    \includegraphics[width=\linewidth]{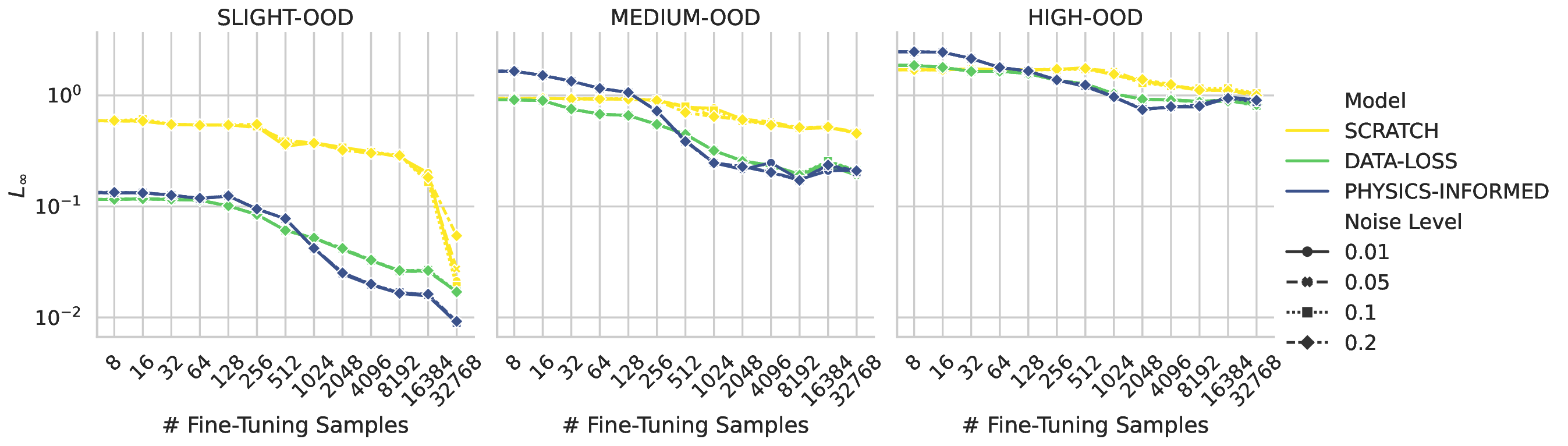}
    \caption{Evaluation of models fine-tuned with varying numbers of noisy downstream samples for the n-shot Advection-Diffusion task, using the \(L_{\infty}\) metric.}
    \label{appfig:n-shot:robustness:ad:linf}
\end{figure}

\begin{figure}[!htbp]
    \centering
    \includegraphics[width=\linewidth]{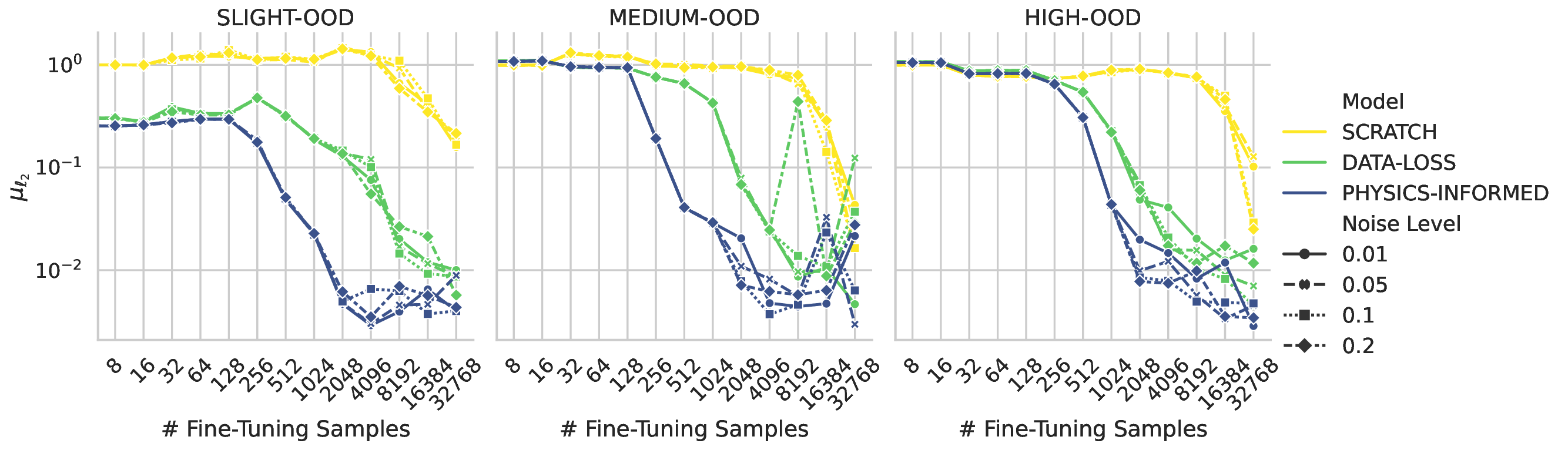}
    \caption{Evaluation of models fine-tuned with varying numbers of noisy downstream samples for the n-shot Helmholtz task, using the \(\mu_{\ell_2}\) metric.}
    \label{appfig:n-shot:robustness:helm:mul2}
\end{figure}

\begin{figure}[!htbp]
    \centering
    \includegraphics[width=\linewidth]{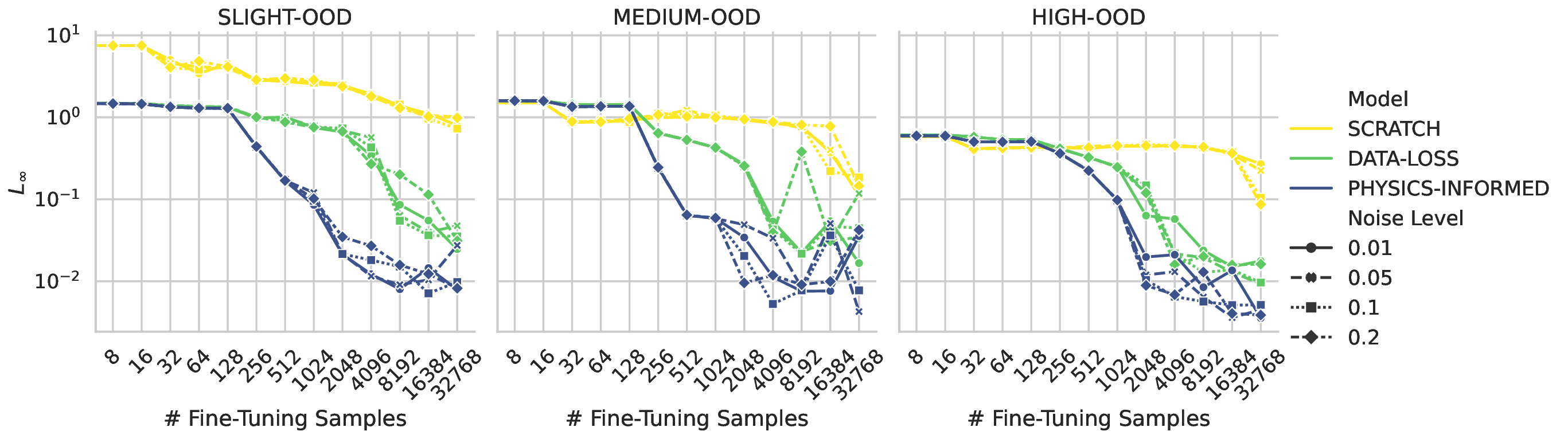}
    \caption{Evaluation of models fine-tuned with varying numbers of noisy downstream samples for the n-shot Helmholtz task, using the \(L_{\infty}\) metric.}
    \label{appfig:n-shot:robustness:helm:linf}
\end{figure}

\begin{figure}[!htbp]
    \centering
    \includegraphics[width=\linewidth]{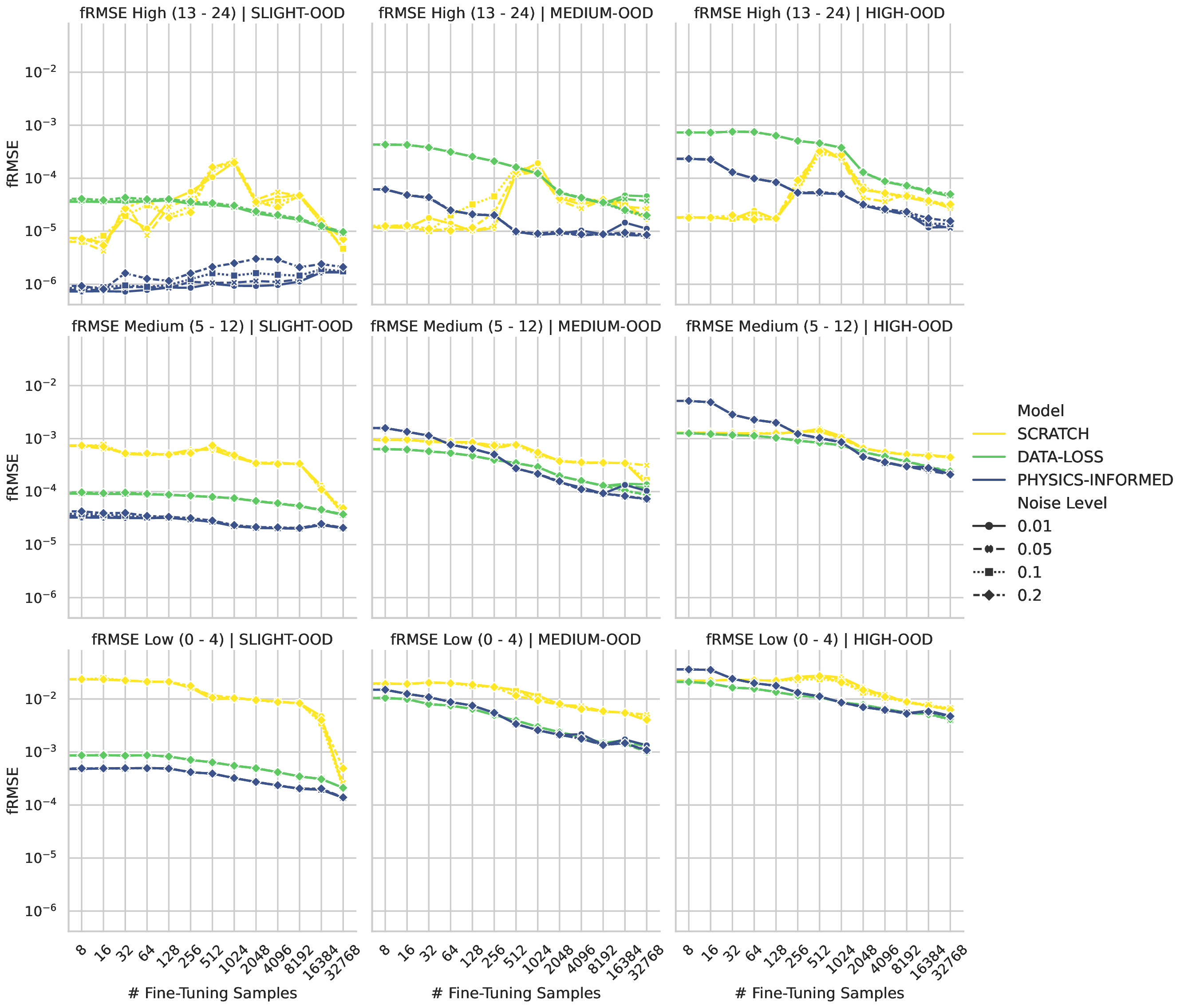}
    \caption{Evaluation of models fine-tuned with varying numbers of noisy downstream samples for the n-shot Advection-Diffusion task, using the fRMSE metric.}
    \label{appfig:n-shot:robustness:ad:frmse}
\end{figure}

\begin{figure}[!htbp]
    \centering
    \includegraphics[width=\linewidth]{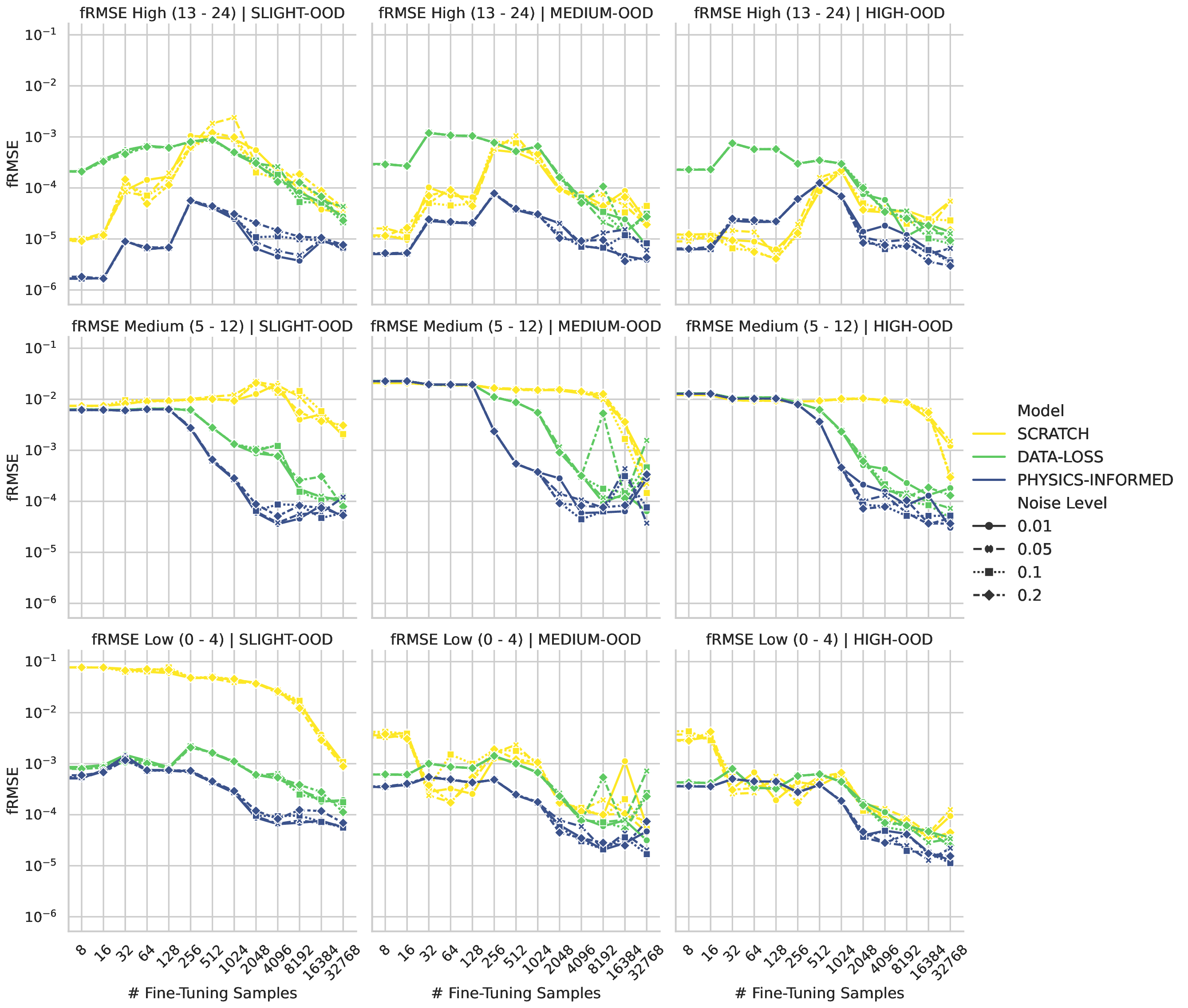}
    \caption{Evaluation of models fine-tuned with varying numbers of noisy downstream samples for the n-shot Helmholtz task, using the fRMSE metric.}
    \label{appfig:n-shot:robustness:helm:frmse}
\end{figure}

\paragraph{Advection-Diffusion.}
\cref{appfig:n-shot:robustness:ad:mul2,appfig:n-shot:robustness:ad:linf} show the \(\mu_{\ell_2}\) and \(L_{\infty}\) metrics for the Advection-Diffusion task under varying noise levels in fine-tuning data. Compared to the noise-free setting, Physics-Informed models remain largely unaffected by noise and consistently outperform baseline models across all noise levels.
\paragraph{Helmholtz.}
\cref{appfig:n-shot:robustness:helm:mul2,appfig:n-shot:robustness:helm:linf} show the \(\mu_{\ell_2}\) and \(L_{\infty}\) metrics for the Helmholtz task under varying noise levels in the downstream fine-tuning data. In low-data regimes, performance in out-of-distribution settings remains relatively stable, mirroring trends observed for the Advection-Diffusion task. In contrast, high-data regimes (e.g., 2048 downstream samples) exhibit increased performance fluctuations with no clear pattern.
The trend observed in the Helmholtz task suggests that incorporating the \(\mathcal{L}_{\text{PDE}}\) signal during pre-training leads to more robust downstream models. Since Additive White Gaussian Noise predominantly affects high-frequency components, the fRMSE metric at higher frequencies serves as a more sensitive indicator of model performance under noisy conditions. The fRMSE results for the Advection-Diffusion and Helmholtz tasks are shown in \cref{appfig:n-shot:robustness:ad:frmse,appfig:n-shot:robustness:helm:frmse}, where Physics-Informed models consistently achieve lower fRMSE values in high-frequency domains, even when fine-tuned on noisy data, mirroring behavior in the noise-free n-shot setting. This frequency-aware robustness highlights the benefits of \(\mathcal{L}_{\text{PDE}}\), which helps the model capture small-scale, high-frequency features that are typically missed by purely data-driven methods. By aligning more closely with the true frequency characteristics of the underlying PDEs, these models exhibit enhanced resilience to high-frequency noise.
\subsection{New Operators}

\subsubsection{Reaction-Diffusion Analysis}
\label{sec:new-operators:reaction-diffusion}
\begin{figure}[!htbp]
    \centering
    \includegraphics[width=0.8\linewidth]{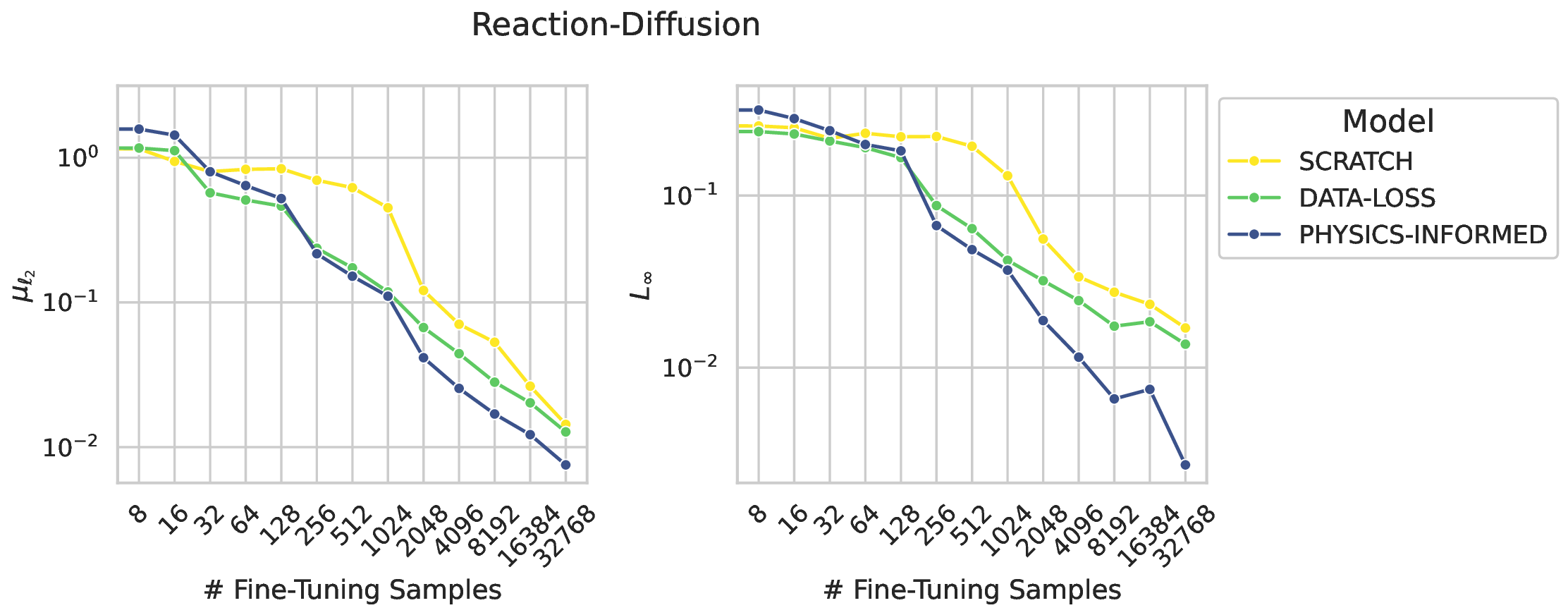}
    \caption{Evaluation of models fine-tuned with increasing numbers of downstream samples for the n-shot Reaction-Diffusion task.}
    \label{appfig:new-operators:rd:exp}
\end{figure}

\begin{figure}[!htbp]
    \centering
    \includegraphics[width=1\linewidth]{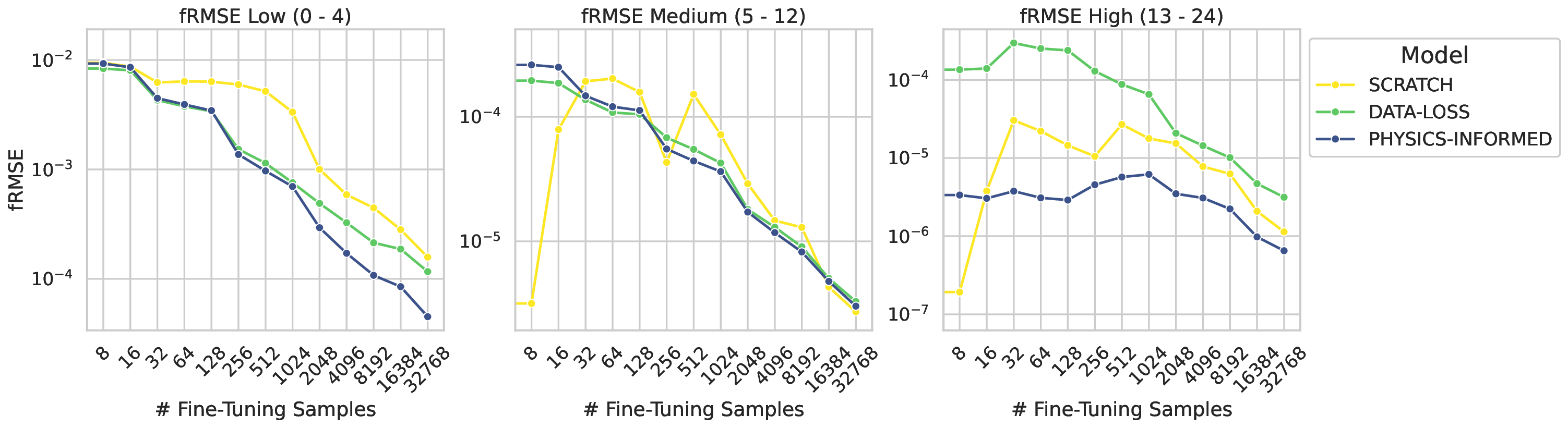}
    \caption{The fRMSE for low frequencies, medium frequencies, and high frequencies for the downstream tasks of the Reaction-Diffusion problem.}
    \label{appfig:new-operators:rd:frmse}
\end{figure}

\cref{appfig:new-operators:rd:exp} shows the results for models fine-tuned on the downstream Reaction-Diffusion task. In low-data regimes, all models, including the pre-trained ones, struggle and perform comparably to the Scratch model, indicating limited ability to learn the task under scarce supervision. Only once fine-tuned on 256 or more downstream samples do the pre-trained models improve over the Scratch model, demonstrating their ability to leverage prior knowledge once sufficient supervision is available. As shown in \cref{appfig:new-operators:rd:frmse}, they begin to capture the dominant low-frequency components from this point onward. This performance gap persists in higher data regimes, underscoring the benefit of pre-training for generalization to unseen systems. While the Physics-Informed model achieves lower errors with fewer samples compared to the purely data-driven (Data-Loss) variant, its improvement over the baseline pre-trained model is only moderate. This suggests that, while the physics signal adds some benefit, most of the generalization gain stems from the pre-training itself.

\subsubsection{Reaction-Advection-Diffusion Analysis}
\label{sec:new-operators:reaction-advection-diffusion}
\begin{figure}[!htbp]
    \centering
    \includegraphics[width=0.8\linewidth]{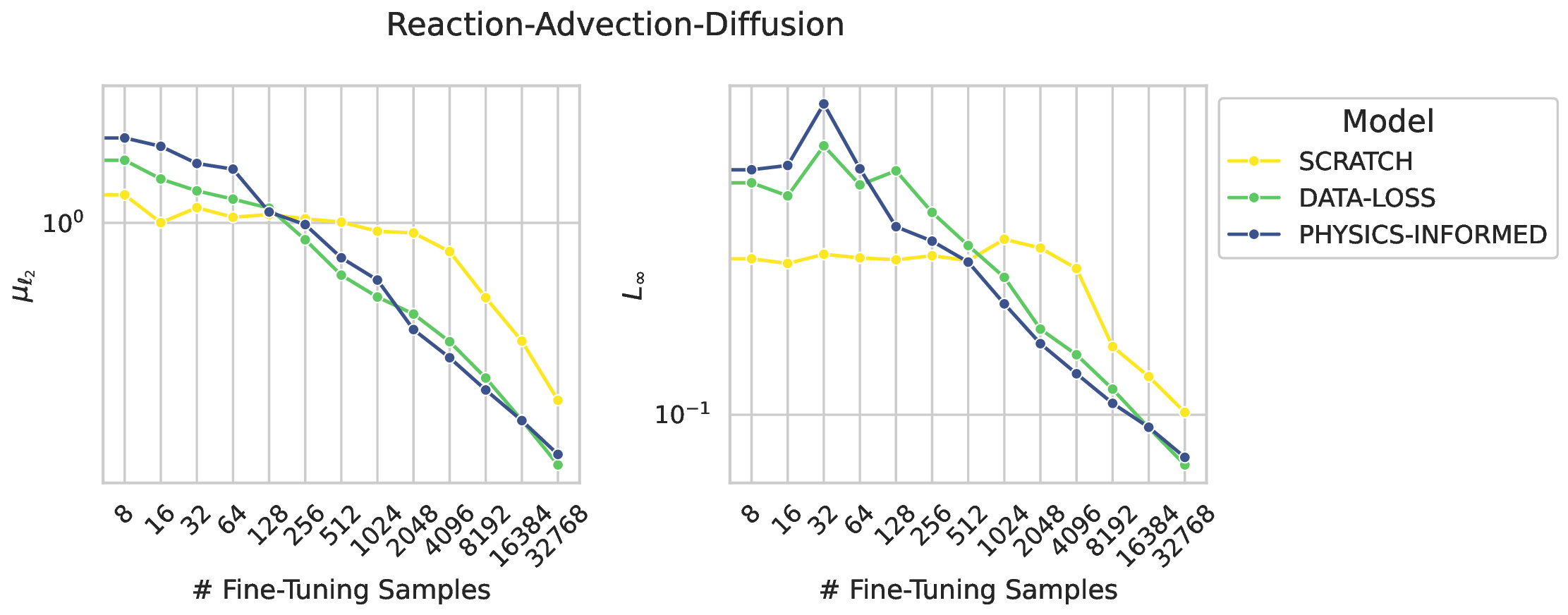}
    \caption{Evaluation of models fine-tuned with increasing numbers of downstream samples for the n-shot Reaction-Advection-Diffusion task.}
    \label{appfig:new-operators:rad:exp}
\end{figure}

\begin{figure}[!htbp]
    \centering
    \includegraphics[width=\linewidth]{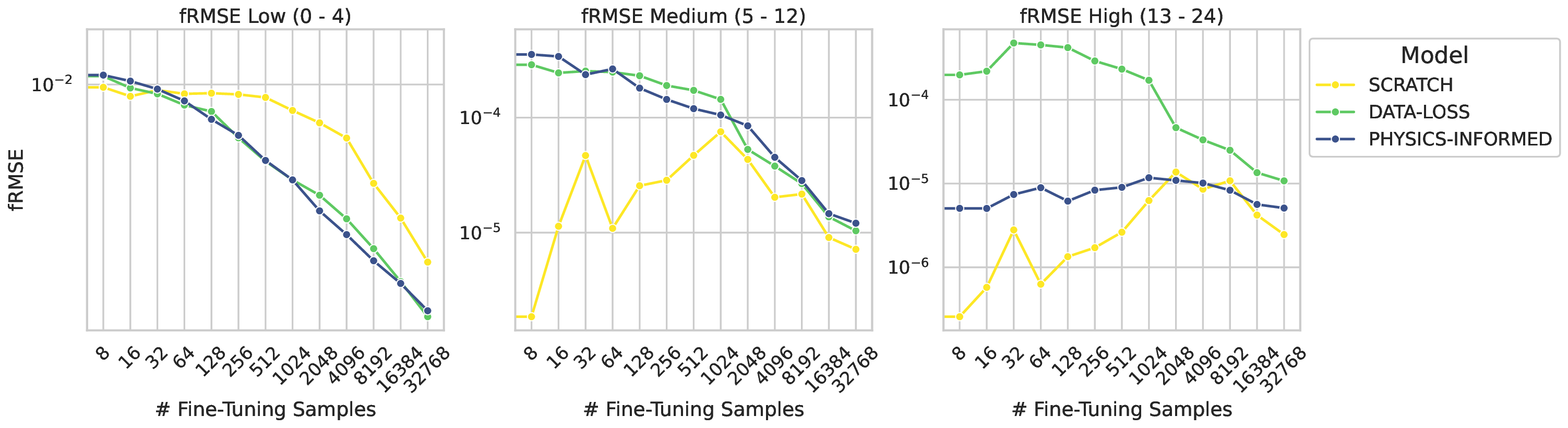}
    \caption{The fRMSE for low frequencies, medium frequencies, and high frequencies for the downstream tasks of the Reaction-Advection-Diffusion problem.}
    \label{appfig:new-operators:rad:frmse}
\end{figure}

While the Reaction-Advection-Diffusion task does not introduce any new coefficients over the Reaction-Diffusion task, as the advection process is present in the pre-training data, the inclusion of the additional velocity field represented by the vector \(\mathbf{v}\) significantly alters the underlying solution space, resulting in a more complex system. This increase in complexity is analogous to the transition observed from the Poisson task to the Advection-Diffusion task, where the introduction of advection dynamics necessitated a better representation from the model to capture the interactions within the system. This shift is evident in the results presented in \cref{appfig:new-operators:rad:exp}. We observe trends similar to those seen in challenging out-of-distribution tasks for Helmholtz and Advection-Diffusion. Specifically, the Physics-Informed model struggles to significantly reduce error rates compared to the baseline Data-Loss model in these scenarios, highlighting the sensitivity of the residual-based training signal to significant changes in the solution space. This can be attributed to the model's inability to learn the dominant low-frequency components, as shown in~\cref{appfig:new-operators:rad:frmse}. This observation is consistent with findings from both the zero-shot and n-shot analyses, reinforcing the notion that, while physics-informed approaches can be beneficial, they may also be vulnerable to substantial changes in the underlying dynamics of the system.

\subsubsection{Navier-Stokes Analysis (Laminar)}
\label{sec:new-operators:navier-stokes}
\begin{figure}[!htbp]
    \centering
    \includegraphics[width=0.8\linewidth]{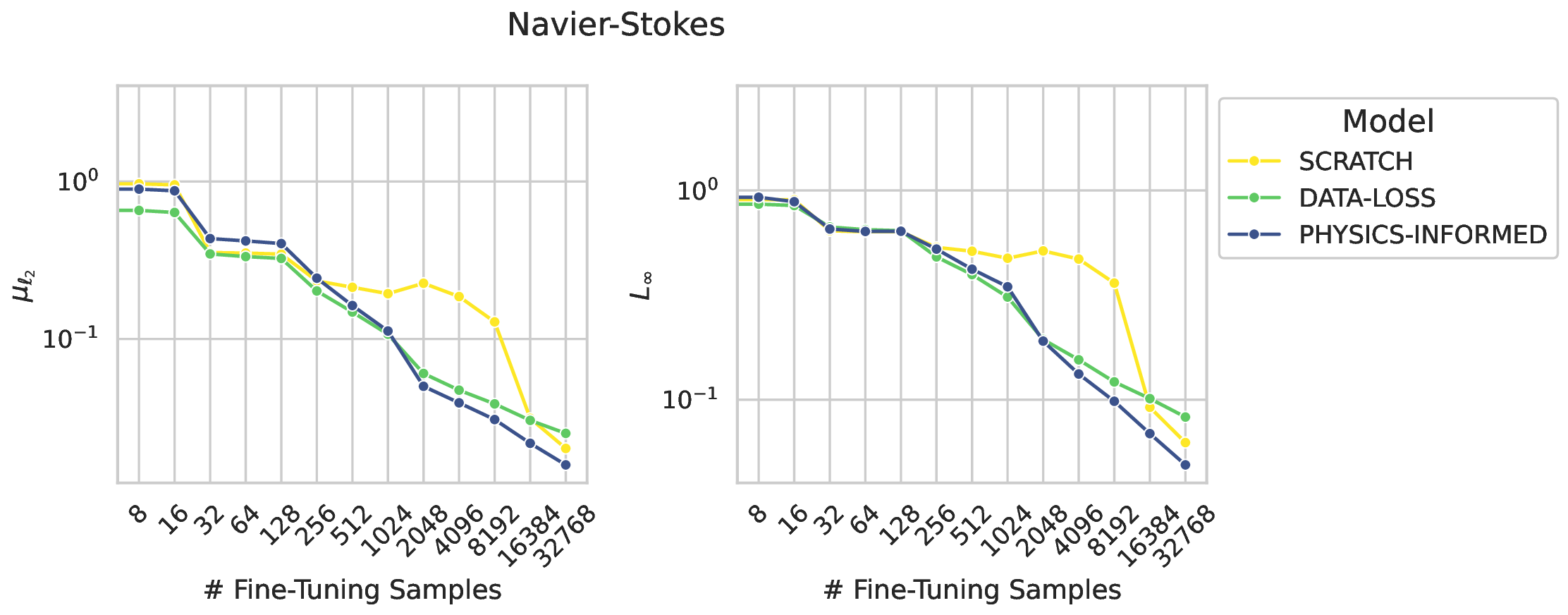}
    \caption{Evaluation of models fine-tuned with increasing numbers of downstream samples for the n-shot Navier-Stokes task.}
    \label{appfig:new-operators:nav:exp}
\end{figure}

\begin{figure}[!htbp]
    \centering
    \includegraphics[width=\linewidth]{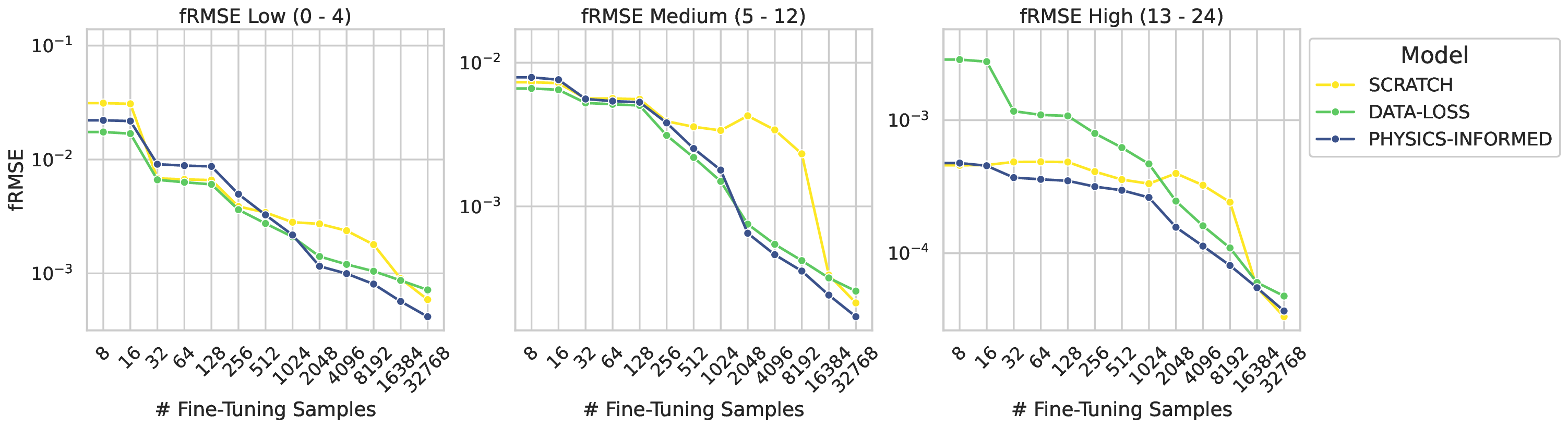}
    \caption{The fRMSE for low frequencies, medium frequencies, and high frequencies for the downstream tasks of the Navier-Stokes (Laminar) problem.}
    \label{appfig:new-operators:ns-laminar:frmse}
\end{figure}

The Navier-Stokes equation introduces a new dynamic through the kinematic viscosity coefficient \(\nu\), fundamentally altering the system compared to the pre-training tasks. Moreover, it transitions the problem from the steady-state formulations used in pre-training to an initial value problem, adding temporal complexity and making it more challenging for models to learn and generalize. We observe that both pre-trained models require 512 or more fine-tuning samples before adapting effectively and reducing errors below the Scratch baseline. However, the Physics-Informed model (\cref{appfig:new-operators:nav:exp}) does not outperform its data-driven counterpart, suggesting that in this setting, the inclusion of a soft physics constraint offers limited additional benefit. Both pre-trained models outperform the Scratch model in capturing medium-frequency components. However, as shown in~\cref{appfig:new-operators:ns-laminar:frmse}, the dominant frequencies lie in the lower range, where all models exhibit similar performance.

\subsubsection{Darcy Analysis}
\label{sec:new-operators:darcy}
The Darcy task analyzes fluid flow through porous media, governed by the spatially varying permeability tensor \(\mathbf{K}\). Unlike the scalar coefficients used in pre-training tasks, \(\mathbf{K}\) introduces anisotropy and heterogeneity into the system, significantly increasing its complexity. Additionally, the Darcy task imposes Dirichlet boundary conditions, fixed values at the domain boundaries, which marks a notable shift from the periodic boundary conditions used during pre-training.

\begin{figure}[!htbp]
    \centering
    \includegraphics[width=0.8\linewidth]{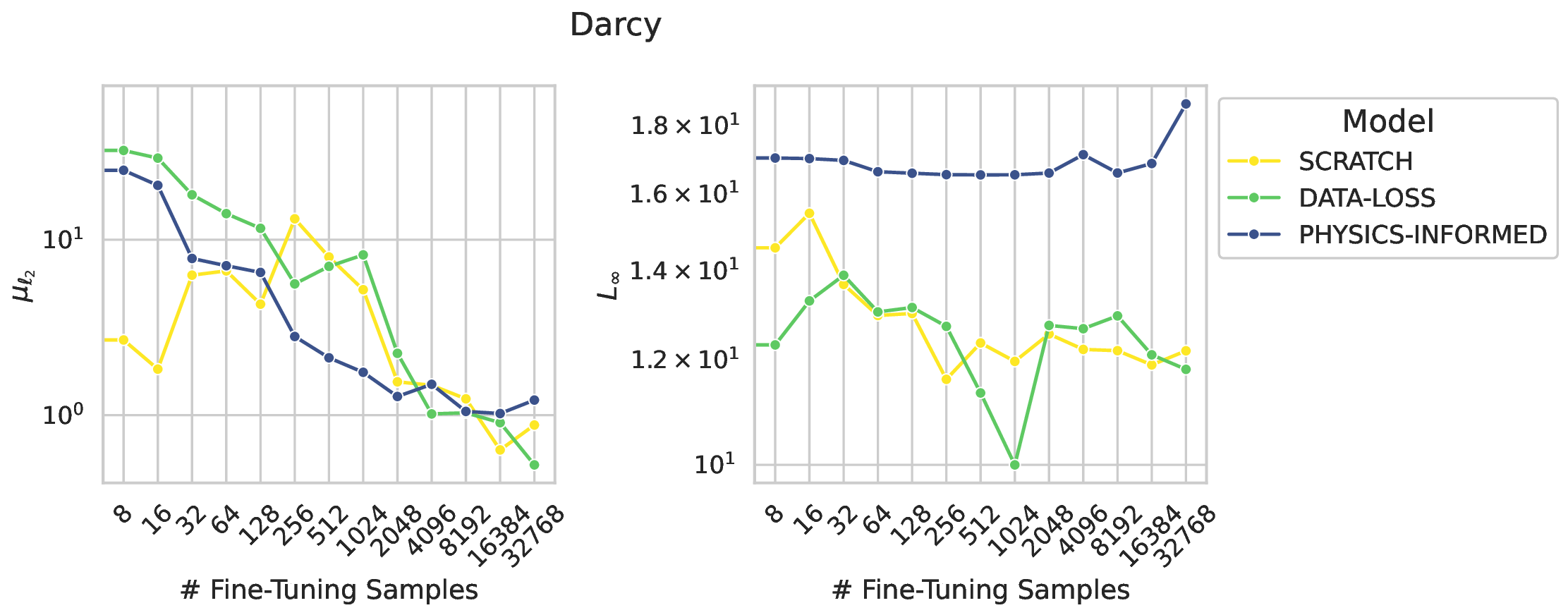}
    \caption{Evaluation of models fine-tuned with increasing numbers of downstream samples for the n-shot Darcy task.}
    \label{appfig:new-operators:darcy}
\end{figure}

\begin{figure}[!htbp]
    \centering
    \includegraphics[width=\linewidth]{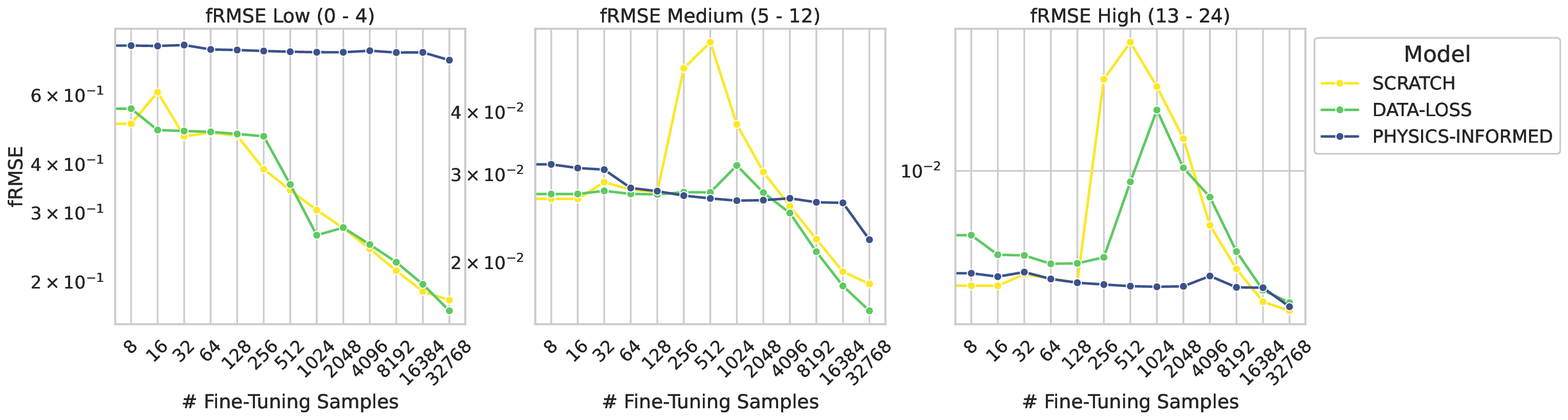}
    \caption{The fRMSE for low frequencies, medium frequencies, and high frequencies for the downstream tasks of the Darcy flow problem.}
    \label{appfig:new-operators:darcy:frmse}
\end{figure}

In contrast to previous new-operator tasks, all models, regardless of loss function or pre-training strategy, fail to approximate the Darcy operator effectively. This failure is most pronounced in the Physics-Informed model, which consistently performs the worst, as shown in \cref{appfig:new-operators:darcy}. While the spatially varying coefficient tensor likely contributes to the poor performance, the shift in boundary conditions appears to pose an even greater challenge. Dirichlet boundaries lead to solutions with non-periodic behavior, in stark contrast to the structure of the pre-training PDEs. Although finite difference methods are used to compute derivatives for \(\mathcal{L}_{\text{PDE}}\), since spectral derivatives are inaccurate for non-periodic problems, the Physics-Informed model struggles to capture the new frequency components present in the solution. This is likely due to overfitting to the dominant frequency modes of the pre-training data (as shown in~\cref{appfig:new-operators:darcy:frmse}), highlighting a key limitation of the current physics-informed approach when applied to systems with mismatched boundary and frequency characteristics.

\end{document}